\documentclass[lettersize,journal]{IEEEtran}
\usepackage{amsmath,amsfonts}
\usepackage{algorithmic}
\usepackage{algorithm}
\usepackage{array}
\usepackage[caption=false,font=footnotesize,labelfont=sf,textfont=sf]{subfig}
\usepackage{textcomp}
\usepackage{subfig}
\usepackage{stfloats}
\usepackage{url}
\usepackage{verbatim}
\usepackage{graphicx}
\usepackage{cite}
\usepackage{tikz}
\usetikzlibrary{shapes.geometric, arrows, decorations.pathreplacing}
\usetikzlibrary{backgrounds}
\usepackage{xcolor}
\usepackage{adjustbox}
\usepackage{wrapfig}
\usepackage{wrapfig}
\usepackage{booktabs}
\usepackage{pgfplots}
\usepackage{hyperref}
\usepackage{makecell}
\usepackage{multirow}
\pgfplotsset{compat=1.18}

\begin{document}
\definecolor{network-blue}{RGB}{197, 223, 248}
\definecolor{plan-emb-c}{RGB}{229,239,193}
\definecolor{language-goal-emb-c}{RGB}{162,213,171}
\definecolor{static-obs-emb-c}{RGB}{57,174,169}
\definecolor{gripper-obs-emb-c}{RGB}{85,123,131}
\definecolor{language-emb-c}{RGB}{204,214,166}
\definecolor{light-yellow}{RGB}{238, 233, 218}
\definecolor{light-green}{RGB}{211, 235, 205}
\definecolor{light-red}{RGB}{242, 182, 160}
\definecolor{light-blue}{RGB}{127, 161, 195}
\definecolor{environment-c}{RGB}{254, 242, 244}
\definecolor{emb-c}{RGB}{100, 130, 173}
\definecolor{gt-c}{RGB}{155, 164, 181}
\definecolor{dark-red}{RGB}{188, 71, 73}
\definecolor{dark-green}{RGB}{56, 102, 65}
\definecolor{standard-gaussian}{RGB}{230, 226, 195}
\definecolor{prior-distribution}{RGB}{136, 164, 124}
\definecolor{skill-instance}{RGB}{28, 49, 94}
\title{Language-Conditioned Imitation Learning with Base Skill Priors under Unstructured Data}

\author{Hongkuan Zhou$^{1}$, Zhenshan Bing $^{1\,\dagger}$, Xiangtong Yao$^{1}$, \\ Xiaojie Su$^{2}$, Chenguang Yang$^3$, Kai Huang$^4$, Alois Knoll$^1$ % <-this % stops a space
\thanks{$^1$ Techinical University of Munich, Munich, Germany}% <-this % stops a space
\thanks{$^2$ Chongqing University, ChongQing, China}
\thanks{$^3$ Department of Computer Science, University of Liverpool, U.K.}
\thanks{$^4$ Sun Yat-sen University, Guangzhou, China}
\thanks{$^{\dagger}$Corresponding author: Zhenshan Bing {\tt\small zhenshan.bing@tum.de}}
}



\maketitle
\pagestyle{empty}
\begin{abstract}
The growing interest in language-conditioned robot manipulation aims to develop robots capable of understanding and executing complex tasks, with the objective of enabling robots to interpret language commands and manipulate objects accordingly. While language-conditioned approaches demonstrate impressive capabilities for addressing tasks in familiar environments, they encounter limitations in adapting to unfamiliar environment settings. In this study, we propose a general-purpose, language-conditioned approach that combines base skill priors and imitation learning under unstructured data to enhance the algorithm's generalization in adapting to unfamiliar environments. We assess our model's performance in both simulated and real-world environments using a zero-shot setting. 
The average completed task length, indicating the average number of tasks the agent can continuously complete, improves more than 2.5 times compared to the baseline method HULC. In terms of the zero-shot evaluation of our policy in a real-world setting, we set up ten tasks and achieved an average 30\% improvement in our approach compared to the current state-of-the-art approach, demonstrating a high generalization capability in both simulated environments and the real world. For further details, including access to our appendix, code base, and videos, please refer to this link \url{https://hk-zh.github.io/spil/}.
\end{abstract}


\section{Introduction}\label{section:introduction}

Language-conditioned robot manipulation \cite{zhou2024languageconditionedlearningroboticmanipulation} is an emerging field of research at the intersection of robotics, natural language processing, and computer vision. This domain seeks to develop robots capable of understanding their surrounding environments and executing complex manipulation tasks based on natural language commands provided by humans. Substantial progress has been made in recent years, with some studies focusing on deep reinforcement learning techniques to shape reward functions for language instructions, enabling agents to solve tasks through trial-and-error processes by following language instructions \cite{DBLP:journals/corr/abs-1806-01946, pmlr-v164-nair22a, pmlr-v155-goyal21a, bing23million, 10341769}. 
They are well-designed to address the low sample efficiency and enable effective learning. 
Other researchers also leverage language-conditioned imitation learning approaches, which train agents using demonstration datasets.
For instance, some studies utilize imitation learning with expert demonstrations that are accompanied by labeled language instructions to solve such language-conditioned tasks \cite{stepputtis2020language, jang2021bcz}. While these methods have demonstrated a high success rate in completing tasks, two main shortcomings still exist. Firstly, the process is limited by the substantial effort required to sample expert demonstrations. As a result, the dataset available for exploring various scenarios in the environment is restricted, ultimately hindering the agent's potential for better performance. Secondly, the trained agent is deficient in its capacity for generalization, which impedes its ability to carry out tasks in unseen environments. 

\tikzstyle{arrow} = [thick,->,>=stealth]
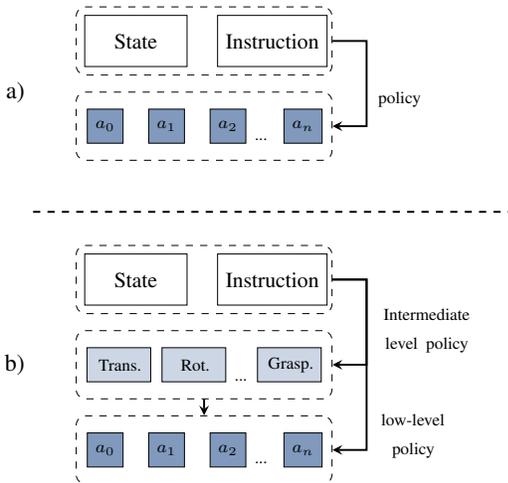
\begin{figure}[t]
    \centering
    \begin{adjustbox}{width=0.45\textwidth}
        \begin{tikzpicture}[node distance=2cm]
            % \node (a) [rectangle, minimum height=3.5cm, minimum width=7cm, label={[label distance=0cm]-180: a)}] at (0,2){};
            % \node (state_a) [rectangle, draw=black, minimum height=0.75cm, minimum width=1.5cm, fill=white, above of=a, xshift=-2cm, yshift=-1.25cm]{\footnotesize State};
            % \node (instruction_a) [rectangle, draw=black, minimum height=0.75cm, minimum width=1.5cm, fill=white, right of=state_a, xshift=0cm]{\footnotesize Instruction};
            % \node (border1_a) [rectangle, rounded corners, dashed, draw=black, minimum height=1cm, minimum width=3.75cm, above of=a, yshift=-1.25cm, xshift=-1cm]{};
            % \node (action0_a) [rectangle, draw=black, minimum height=0.5cm, minimum width=0.5cm, fill=emb-c!50, below of=state_a, xshift=-0.45cm, yshift=0.75cm]{\scriptsize $a_0$};
            % \node (action1_a) [rectangle, draw=black, minimum height=0.5cm, minimum width=0.5cm, fill=emb-c!50, right of=action0_a, xshift=-1.1cm]{\scriptsize $a_1$};
            % \node (action2_a) [rectangle, draw=black, minimum height=0.5cm, minimum width=0.5cm, fill=emb-c!50, right of=action1_a, xshift=-1.1cm,label={[label distance=0cm]-10: \scriptsize ...}]{\scriptsize $a_2$};
            % \node (actionn_a) [rectangle, draw=black, minimum height=0.5cm, minimum width=0.5cm, fill=emb-c!50, right of=action2_a, xshift=-0.9cm]{\scriptsize $a_n$};
            % \node (border2_a) [rectangle, rounded corners, dashed, draw=black, minimum height=1cm, minimum width=3.75cm, below of=border1_a, yshift=0.75cm]{};
            % \draw [arrow] (border1_a.east) -- +(0.5, 0) |- (border2_a.east) node[right,midway, text width=1.5cm, text centered, yshift=0.4cm, xshift=-0.4cm] {\scriptsize policy};
                        
            \node (b) [rectangle, minimum height=4cm, minimum width=7cm] at (0,0){};
            \node (state_b) [rectangle, draw=black, minimum height=0.75cm, minimum width=1.5cm, fill=white, above of=b, xshift=-2cm, yshift=-0.75cm]{\footnotesize State};
            \node (instruction_b) [rectangle, draw=black, minimum height=0.75cm, minimum width=1.5cm, fill=white, right of=state_b, xshift=0cm]{\footnotesize Instruction};
            \node (border1_b) [rectangle, rounded corners, dashed, draw=black, minimum height=1cm, minimum width=3.75cm, above of=b, yshift=-0.75cm, xshift=-1cm]{};
            \node (skill0_b) [rectangle, draw=black, minimum height=0.5cm, minimum width=0.65cm, text width=0.7cm, text centered, fill=emb-c!50, below of=state_b, xshift=-0.25cm, yshift=0.75cm]{\scriptsize Trans.};
            \node (skill1_b) [rectangle, draw=black, minimum height=0.5cm, minimum width=0.65cm, text width=0.7cm, text centered, fill=emb-c!50, right of=skill0_b, xshift=-0.9cm, label={[label distance=0cm]-10: \scriptsize ...}]{\scriptsize Rot.};
            \node (skilln_b) [rectangle, draw=black, minimum height=0.5cm, minimum width=0.65cm, fill=emb-c!50, text width=0.7cm, text centered, right of=skill0_b, xshift=0.5cm]{\scriptsize Grasp.};
            \node (border2_b) [rectangle, rounded corners, dashed, draw=black, minimum height=1cm, minimum width=3.75cm, below of=border1_b, yshift=0.75cm]{};
            \node (action0_b) [rectangle, draw=black, minimum height=0.5cm, minimum width=0.5cm, fill=emb-c, below of=skill0_b, xshift=-0.2cm, yshift=0.75cm]{\scriptsize $a_0$};
            \node (action1_b) [rectangle, draw=black, minimum height=0.5cm, minimum width=0.5cm, fill=emb-c, right of=action0_b, xshift=-1.1cm]{\scriptsize $a_1$};
            \node (action2_b) [rectangle, draw=black, minimum height=0.5cm, minimum width=0.5cm, fill=emb-c, right of=action1_b, xshift=-1.1cm,label={[label distance=0cm]-10: \scriptsize ...}]{\scriptsize $a_2$};
            \node (actionn_b) [rectangle, draw=black, minimum height=0.5cm, minimum width=0.5cm, fill=emb-c, right of=action2_b, xshift=-0.9cm]{\scriptsize $a_n$};
            \node (border3_b) [rectangle, rounded corners, dashed, draw=black, minimum height=1cm, minimum width=3.75cm, below of=border2_b, yshift=0.75cm]{};

            \draw [arrow, draw=dark-green] (border1_b.east) -- +(0.5, 0) |- (border2_b.east) node[right,midway, text width=1.5cm, text centered, yshift=0.5cm] {\scriptsize Intermediate level policy};
            \draw [arrow,draw=dark-red, dashed] (border1_b.west) -- +(-0.5, 0) |- (border3_b.west)node[left,midway, text width=1.5cm, text centered, yshift=1.25cm, xshift=0.25cm] {\scriptsize policy};
            \draw [arrow, draw=dark-green] (border1_b.east) -- +(0.5, 0) |- (border3_b.east) node[right,midway, text width=1.5cm, text centered, yshift=0.2cm, xshift=-0.2cm] {\scriptsize low-level policy};
            \draw [arrow] (border2_b.south) -- (border3_b.north);
        \end{tikzpicture}
    \end{adjustbox}
    \caption{\small Comparison of common approaches (\textcolor{dark-red}{dashed red}) and our approach (\textcolor{dark-green}{green}). Common approaches usually directly learn the actions, depending on current observation and instruction. Our approach aims to learn the extra intermediate-level policy of which base skill to choose, based on current observation and instruction. }
    \label{fig:intermedia-policy}
    \vspace{-1em}
\end{figure}

To address the first problem, some researchers employ unstructured data (play data) \cite{lynch2019play, lynch2021language, rosetebeas2022latentplanstaskagnosticoffline, cui2022play, borja2022affordance},  which consists of human demonstrations driven by curiosity or other intrinsic motivations, rather than being driven by specific tasks, to reduce the effort required to collect expert data for training. 
All the play data is obtained through interactions with simulation environments by participants using virtual reality (VR) equipment, with only 1 \% of the data is labeled with language instruction.
% A substantial 99\% of the training data is comprised of play data, obtained through interactions with simulation environments by participants using virtual reality (VR) equipment. Only 1\% of the data is labeled with language instructions. 
By employing play data, the labor-intensive task of data labeling is significantly reduced, facilitating the creation of larger training datasets for imitation learning. The trained agent demonstrates remarkable performance, exhibiting a high success rate across various tasks. Building upon the ideas presented in \cite{lynch2021language}, HULC \cite{9849097} was developed to enhance the performance of language-conditioned imitation learning by integrating transformer structures and contrastive representation learning. 
HULC++ \cite{mees2023grounding} further improves the performance by incorporating  a self-supervised visuo-lingual affordance model.

Regarding the second problem, current approaches still face a challenge in generalizing to perform tasks in unfamiliar and complex environments. The policy learned through the imitation learning algorithm exhibits outstanding evaluation performance, primarily in training domains, suggesting that the policy's effectiveness is restricted to scenarios where training and evaluation environments are identical. Upon conducting sim2real experiments and zero-shot evaluations in novel environments, the discrepancy between the evaluation and training environments results in a substantial decline in success rates. 
% The Zero-shot Multi-environment evaluation provided by the CALVIN benchmark \cite{mees2022calvin} further highlights the limitations of current language-conditioned imitation learning approaches (the success rate decreased by 50 percent), illustrating that trained agents struggle to ground language instructions to target objects and actions in unfamiliar environments. 

Within the imitation learning framework, agents typically rely on predicting the short-term next action at each time step based on the current observation and goal without learning a high-level long-term procedure. This approach diverges from the more natural approach employed by humans, which typically involves breaking down complex tasks into simpler, basic steps. Skill-based learning \cite{shiskill, nagabandi2020deep} is a promising approach that utilizes pre-defined skills to expedite the learning process, leveraging the prior knowledge encoded within these skills, which is typically derived from human expertise. A primary factor contributing to the suboptimal performance of current language-conditioned imitation learning methodologies is the absence of prior knowledge during the training process. The excessive dependence on training data can lead to overfitting and impede generalization to unfamiliar scenarios. By incorporating prior skills into the learning process, the agent can avoid the necessity to start from scratch and reduce the dependency of training data.

In this paper, we introduce a base \textbf{S}kill \textbf{P}rior based \textbf{I}mitation \textbf{L}earning (\textbf{SPIL}) framework designed to enhance the generalization ability of an agent in adapting to unfamiliar environments by integrating base skill priors: translation, rotation, and grasping. Specifically, SPIL learns both a low-level policy for skill instance execution based on observations, as well as an intermediate-level policy that determines which base skill (translation, rotation, and grasping) should be performed under the current observation. Figure \ref{fig:intermedia-policy} compares our approach with normal approaches. The intermediate-level policy functions as a manager, interpreting language instructions and appropriately combining these base skills to solve manipulation tasks. For instance, when the intermediate-level policy receives the language instruction ``lift the block'', it will decompose the task into several steps involving base skills, such as approaching the block (translation), grasping the block (grasping), and lifting the block (translation). Note that the reason we call it intermediate-level policy is to distinguish it from the more complex high-level policy for tasks like ``tidying up the room'' which can be decomposed into several subtasks (usually done by LLMs \cite{wu2023tidybot}). We evaluate our algorithm using the CALVIN benchmark \cite{mees2022calvin} and achieve outstanding performance in the challenging zero-shot multi-environment settings. Furthermore, we conduct sim-to-real experiments to assess the performance of our approach in real-world environments, yielding outstanding results. We summarize the key contributions as follows: 
\begin{itemize}
    \item In this paper, we incorporate the skill priors into imitation learning and design a skill-prior-based imitation learning mechanism to enable learning of an intermediate-level procedure and enhance the generalization ability of the learned policy.
    \item Our proposed method exhibits superior performance compared to previous baselines, particularly in terms of its ability to generalize and perform well in previously unseen environments. Our evaluation shows that our approach outperforms the current method HULC by a significant margin, achieving 2.5 times the performance. We conducted a series of sim-to-real experiments to investigate further our model's generalization ability in unseen environments and the potential of our model for real-world applications.
\end{itemize}

\section{Related works}\label{section:related_works}
% Grounding language to behaviors based on vision observations has attracted significant interest and attention within the field of robotics. 
% This section covers the popular language-robot manipulation framework. 
% Additionally, the skill-based learning methods which inspired our approach are also discussed. 

In the field of language-conditioned robot manipulation, some studies establish connections between visual perception and linguistic comprehension in the vision-and-language field, facilitating the agent's ability to tackle multimodal problems \cite{pont2020connecting, lu2019vilbert, li-etal-2020-bert-vision}. Other research focuses on grounding language instructions and the agent's behaviors, empowering the agent to comprehend instructions and effectively interact with the environment \cite{shridhar2020ingress, 8752407, liu2022structformer, shridhar2022cliport}. However, these approaches employ two-stream architectural models to process multimodal data. Such a model require distinct feature representations for each data modality, such as semantic and spatial representations \cite{shridhar2022cliport},
thus potentially compromising learning efficiency. As an alternative, end-to-end models focus on learning feature representations and decision-making directly from raw input data, where the language instructions as a conditioning factor to train the agent. This approach eliminates the need for manual feature engineering \cite{9849097, coguiding}, thereby offering a more efficient and robust solution for complex tasks and emerging as a trend in language-conditioned robot manipulations.

For instance, imitation learning with end-to-end models has been applied to solve language-conditioned manipulation tasks using expert demonstrations accompanied by a large number of labeled language instructions \cite{stepputtis2020language}  \cite{jang2021bcz}. These approaches necessitate a substantial amount of labeled and structured demonstration data. By extending the idea of \cite{lynch2019play}, Lynch et al. proposed MCIL \cite{lynch2020language}, which grounds the agent's behavior with language instructions using unlabeled and unstructured demonstration data, reducing data acquisition efforts and achieving more robust performance. HULC \cite{9849097}, as an enhanced version of MCIL, is designed to improve the performance of MCIL even further. It has achieved impressive results in the CALVIN benchmark \cite{mees2022calvin} using the single environment setting. However, when tested in the more challenging Zero-shot Multi Environment setting, where the evaluation environment is not exactly the same as the training environments, HULC's performance drops significantly. These suboptimal results suggest that current language-conditioned imitation learning approaches lack the ability to adapt to unfamiliar environments. More recently, some approaches \cite{black2023zeroshotroboticmanipulationpretrained, li2023vision, wu2023unleashinglargescalevideogenerative, ke20243ddiffuseractorpolicy} leverage rich knowledge in the pre-trained foundation models to enhance the generalization ability in unseen environments.

The concept of skill-based mechanisms in deep reinforcement learning provides valuable insights for enhancing the generalizability of algorithms. Specifically, skill-based reinforcement learning leverages task-agnostic experiences in the form of large datasets to accelerate the learning process \cite{hausman2018learning, merel2018neural, ICML-2019-KipfLDZSGKB, Lee2020Learning}. To extract skills from a large task-agnostic dataset, several approaches \cite{pertsch2021skild, pertsch2020spirl} first learn an embedding space of skills and skill priors from the dataset. Inspired by this, we have developed an imitation learning approach that utilizes certain base skill priors. By employing this method, the agent learns intermediate-level processes (composing these base skills) that aid in task completion, thereby enhancing its ability to generalize across different scenarios.
\section{Methodology}\label{section:methodologies}
% In this section, we first formally define the problem of language-conditioned robot manipulation. Then we demonstrate how to extract the base skill embeddings and priors from offline datasets. Finally, we propose our novel imitation learning mechanism with base skill priors. 

% In this section, we first provide an overview of our approach. Following that, we introduce the details of our skill-prior-based imitation learning, which includes the method to construct a continuous skill embedding space with base skill priors and the mechanism to integrate base skill priors into imitation learning. 

\subsection{Overview}
The key idea of our approach is integrating skills into imitation learning by changing the original action space - Cartesian End Effector space $\mathcal{A} \in \mathbb{R}^7$ into skill space $\mathcal{A}_\textrm{skill} \in \mathbb{R}^{N_h \times 7}$, where $N_h$ indicates the horizon of skills. Note that each skill represents a fixed-length ($N_h$) action sequence in our setting. Also, we intend to integrate the concept of base skills (translation, rotation, grasping) into the learning procedure so that the agent can learn an extra intermediate-level policy to decompose tasks into several base skills. Unlike reinforcement learning, the optimization strategy employed in imitation learning involves minimizing the discrepancy between the predicted actions and the corresponding actions observed in the demonstration data. For this reason, a primary challenge in integrating skill priors into imitation learning is the continuous nature of actions in the demonstration data, which requires modeling the skills as a continuous action space to align with the demonstration actions rather than representing the skills by a finite, discrete set of pre-defined action sequences. 
% In reinforcement learning, the optimization mechanism primarily focuses on maximizing the expected rewards rather than specifying the specific action sequences required to complete a task. Hence, it is possible to leverage finite and discrete skills in reinforcement learning. 

To address the challenges mentioned above, the rest of this section is organized as follows:
\begin{enumerate}
    \item We define three base skills (translation, rotation, grasping) for a robotic arm agent and introduce the method to \textbf{stochastically} label action sequences with base skills.
    \item We introduce our approach to learning continuous skill embedding space, integrating base skill priors into such skill space.
    \item By utilizing a continuous skill space and base skills, we implement an imitation learning algorithm to train the agent to acquire the ability to 1) learn an intermediate-level base skill composition to accomplish the desired task and 2) develop a policy that can determine which specific skill instance to perform based on each observation, as opposed to a single action.
\end{enumerate}
The architecture of our proposed method is illustrated in Figure \ref{fig:spil}.

% Inspired from \cite{pertsch2020spirl} which learn a continuous skill space by leveraging encoder-decoder structure, we also encode the action sequence into skill embedding space and 
% Our proposed methodology involves transforming the original 7 degrees of freedom (7DoF) action space into a skill space. Specifically, at each time step, the agent is required to select a base skill and a corresponding skill instance from the skill space based on the current observation and the language instruction. In order to achieve this, there exists two problems to be solved. How to transform the orginal action space of the agent into a skill space? How to use the learned skill prior 

% We define three base skills and intend to learn continuous skill embedding space integrated with these base skills  by variational autoencoder. 
% The approach we propose shares similarities with the methodology presented in \cite{pertsch2020spirl}. However, there is a notable difference in that our approach employs pre-defined base skill priors, namely translation, rotation, and grasping, as opposed to utilizing the observation to encode skill priors as done in \cite{pertsch2020spirl}.

\subsection{Base Skill Labeling}
This section formally defines three base skills - translation, rotation, and grasping. Since each action sequence can contain multiple base skills, deterministically classifying an action sequence to one of three base skills is not reasonable. Here, we stochastically label each given action sequence $x=(a_0, a_1, ... , a_{N_h-1})$ of length $N_h$ with probability ($p({\textrm{trans.}|x)}, p({\textrm{rot.}|x)}, p({\textrm{grasp.}|x)}$) which indicate the probability of $x$ belongs to these three base skills. For example, the probability of (0.7, 0.2, 0.1) suggests a dominance of translation skill within the given action sequence, a minor presence of rotation skill, and a minimal grasping skill. 
We design a non-learning-based approach to label each action sequence. 
Since the action is defined in the Cartesian EE space, it can be accomplished by assessing the accumulated magnitude of seven degrees of freedom within the temporal dimension of a given horizon $N_h$. The probability of this sequence belonging to translation, rotation, and grasping skills can be defined as follows:
\begin{equation}
    \label{eq:base skill label}
    \begin{split}
        p(y|x) = \frac{w_y \cdot \sum_{i=0}^{N_h-1} |a_i^{y}| }{\sum_{k\in \{\textrm{trans., rot., grasp.}\}} w_k \cdot \sum_{i=0}^{N_h-1} |a_i^{k}|}
    \end{split}\text{,}
\end{equation}
where $y \in \{\textrm{trans., rot., grasp.}\}$ refers to base skills and $a = [t_x,t_y,t_z,r_\alpha,r_\beta, r_\gamma, g]$ with $a^{\textrm{trans.}}=[t_x,t_y,t_z]$, $a^{\textrm{rot.}}=[r_\alpha,r_\beta, r_\gamma]$, and $a^{\textrm{grasp.}}=[g]$, indicating the end effector's displacement, rotation, and gripper control. The ``magic weight'' $w_y$ is introduced to address inconsistencies in scale across different units like meters and degrees. These values act as balancing factors and are determined based on our understanding of the inherent relationships between translation, rotation, and grasping. They reflect the subjective nature of defining translation, rotation, and grasping. Since these classifications may be nuanced and depend on human experience, we've chosen 'magic weight' $w_k$ that reflects a common understanding of how these motions are typically defined. 

\tikzstyle{arrow} = [thick,->,>=stealth]
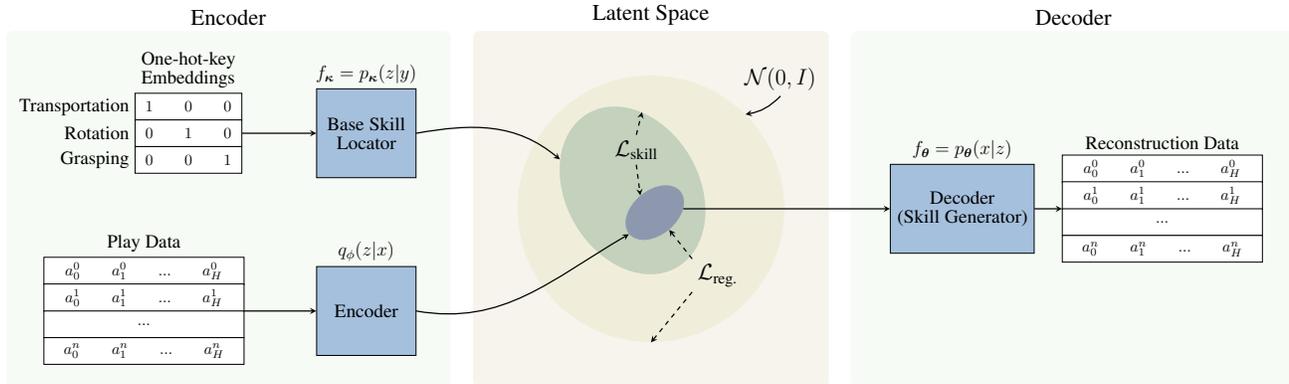
\begin{figure*}[t!]
    \centering
    \begin{adjustbox}{width=0.95\textwidth}
        \begin{tikzpicture}[node distance=2cm]
            \node (encoder-module) [rectangle, rounded corners, fill=light-green!20, minimum height=8cm, minimum width=10cm, label={[label distance=0cm]90:\Large Encoder}] {};
            \node (one-hot-key-t) [rectangle, yshift=2.3cm, xshift=-0.9cm, fill=white, draw=black, minimum height=0.6cm, minimum width=2.4cm, label={[label distance=0cm, xshift=0cm, align=left]90:{\large One-hot-key \\ \large Embeddings}}, label={[label distance=0cm]180:\large Transportation}]{$1 \qquad 0 \qquad  0$};
            \node (one-hot-key-r) [rectangle, below of=one-hot-key-t, yshift=1.4cm, fill=white, draw=black, minimum height=0.6cm, minimum width=2.4cm, label={[label distance=0cm]180:\large Rotation}]{$0 \qquad 1 \qquad  0$};
            \node (one-hot-key-g) [rectangle, below of=one-hot-key-r, yshift=1.4cm, fill=white, draw=black, minimum height=0.6cm, minimum width=2.4cm, label={[label distance=0cm]180:\large Grasping}]{$0 \qquad 0 \qquad  1$};
            \node (play-data-1) [rectangle, below of=one-hot-key-g, yshift=-0.5cm, xshift=-1cm, fill=white, draw=black, minimum height=0.6cm, minimum width=4.5cm, label={[label distance=0cm]90:\large Play Data}]{$a_0^0 \qquad a_1^0 \qquad ... \qquad a_H^0$};
            \node (play-data-2) [rectangle, below of=play-data-1, yshift=1.4cm, fill=white, draw=black, minimum height=0.6cm, minimum width=4.5cm]{$a_0^1 \qquad a_1^1 \qquad ... \qquad a_H^1$};
            \node (play-data-3) [rectangle, below of=play-data-2, yshift=1.4cm, fill=white, draw=black, minimum height=0.6cm, minimum width=4.5cm]{$...$};
            \node (play-data-4) [rectangle, below of=play-data-3, yshift=1.4cm, fill=white, draw=black, minimum height=0.6cm, minimum width=4.5cm]{$a_0^n \qquad a_1^n \qquad ... \qquad a_H^n$};
            \node (skill-prior-encoder) [rectangle, right of=one-hot-key-t, xshift=2cm, yshift=-0.6cm, draw=black, fill=network-blue, minimum height=2cm, minimum width=2cm, text width=2cm, align=center, label={[label distance=0cm]90:\large $f_{\boldsymbol{\kappa}} = p_{\boldsymbol{\kappa}}(z|y)$}]{\large Base Skill Locator };
            \node (encoder) [rectangle, right of=play-data-1, xshift=3cm, yshift=-0.9cm, draw=black, fill=network-blue, minimum height=2cm, minimum width=2cm, text width=2cm, align=center, label={[label distance=0cm]90:\large $q_{\phi}(z|x)$}]{\large Encoder};
            
            \node (latent-space) [rectangle, rounded corners, right of=encoder-module, fill=light-yellow!50, minimum height=8cm, minimum width=8cm, xshift=7.5cm, label={[label distance=0cm]90:\Large Latent Space}] {};
            \node (standard-gaussian) [circle, below of=latent-space, yshift=2cm, fill=standard-gaussian!60, minimum size=6cm] {};
            \node (prior-distribution) [ellipse, above left of=standard-gaussian, yshift=-1cm, xshift=1cm, fill=prior-distribution!50, minimum height=4cm, minimum width=3cm, rotate=30] {};
            \node (skill-instance) [ellipse, below of=prior-distribution, yshift=1.5cm, xshift=0.5cm, fill=skill-instance!50, minimum height=1.5cm, minimum width=1cm, rotate=-50] {};
            \node (label-standard-gaussian) [rectangle, above right of=standard-gaussian, xshift=1.5cm, yshift=1.5cm]{\Large $\mathcal{N}(0, I)$};
            \node (label-reg1) [rectangle, below right of=skill-instance]{\Large $\mathcal{L}_{\textrm{reg.}}$};
            \node (label-reg2) [rectangle, above left of=skill-instance, xshift=1cm]{\Large $\mathcal{L}_{\textrm{skill}}$};
            
            \node (decoder-module) [rectangle, rounded corners,  right of=latent-space, fill=light-green!20, minimum height=8cm, minimum width=10cm, xshift=7.5cm, label={[label distance=0cm]90:\Large Decoder}] {};
            \node (decoder) [rectangle, left of=decoder-module, xshift=-0.5cm, draw=black, fill=network-blue, minimum height=2cm, minimum width=2cm, text width=3cm, align=center, label={[label distance=0cm]90:\large $f_{\boldsymbol{\theta}} = p_{\boldsymbol{\theta}}(x|z)$}]{\large Decoder\\(Skill Generator)};
            
            \node (rec-data-1) [rectangle, right of=decoder-module, yshift=0.9cm, fill=white, draw=black, minimum height=0.6cm, minimum width=4.5cm, label={[label distance=0cm]90:\large Reconstruction Data}]{$a_0^0 \qquad a_1^0 \qquad ... \qquad a_H^0$};
            \node (rec-data-2) [rectangle, below of=rec-data-1, yshift=1.4cm, fill=white, draw=black, minimum height=0.6cm, minimum width=4.5cm]{$a_0^1 \qquad a_1^1 \qquad ... \qquad a_H^1$};
            \node (rec-data-3) [rectangle, below of=rec-data-2, yshift=1.4cm, fill=white, draw=black, minimum height=0.6cm, minimum width=4.5cm]{$...$};
            \node (rec-data-4) [rectangle, below of=rec-data-3, yshift=1.4cm, fill=white, draw=black, minimum height=0.6cm, minimum width=4.5cm]{$a_0^n \qquad a_1^n \qquad ... \qquad a_H^n$};
            
            \draw [arrow] (skill-prior-encoder.west-|one-hot-key-r.east) -- (skill-prior-encoder.west);
            \draw [arrow] (encoder.west-|play-data-2.east) -- (encoder.west);
            \draw [arrow] (skill-prior-encoder.east) to[out=10,in=140] (prior-distribution.north west);
            \draw [arrow] (encoder.east) to[out=-10,in=-150] (skill-instance.south);
            \draw [arrow] (label-standard-gaussian.south) to[out=-135,in=10] (standard-gaussian.north east);
            \draw [arrow, dashed] (label-reg1.south west) -- (standard-gaussian.south);
            \draw [arrow, dashed] (label-reg1.north west) -- (skill-instance.east);
            \draw [arrow, dashed] (label-reg2.north) -- (prior-distribution.north east);
            \draw [arrow, dashed] (label-reg2.south) -- (skill-instance.west);
            \draw [arrow] (skill-instance.north east |- decoder.west)-- (decoder.west);
            \draw [arrow] (decoder.east)-- (decoder.east-|rec-data-1.west);
        \end{tikzpicture}
    \end{adjustbox}
    \caption{\small This architecture comprises two encoders - the action sequence encoder and the base skill locator (encoder), and a decoder for reconstructing the skill embeddings into action sequences. The base skill locator takes one-hot-key embeddings of translation, rotation, and grasping as input and outputs the distribution of the base skill prior in the skill latent space. The action sequence encoder encodes the action sequences with a fixed horizon of $N_h$ to the skill distribution in the latent space. The decoder then reconstructs the skill embedding into action sequences.}
    \label{fig:VAE_Skill_Priors}
    \vspace{-1.5em}
\end{figure*}

\subsection{Continuous Skill Embeddings with Base Skill Priors}
In this section, we introduce a skill space $\mathcal{A}_\textrm{skill} \in \mathbb{R}^{N_h \times 7}$ as the action space for the agent. To better represent such skill space, we compress the action sequences into skill embeddings by following the idea of Variantial AutoEncoders (VAEs), leveraging the action sequences sampled from play data. After training, we acquire a latent space full of skill embeddings and three clusters, indicating the base skills priors for translation, rotation, and grasping. To achieve this, we define $y$ as the variable for base skills and the base skill distribution in the latent space can be written as $z \sim p(z|y)$. For the given action sequence $x$, we employ the approximate variational posterior $q(z|x)$ and $q(y,z|x)$ to estimate the intractable true posterior. Following the VAEs procedure, we measure the Kullback-Leibler (KL) divergence between the true posterior and the posterior approximation to determine the ELBO (the details can be seen in Appendix Theoretical Motivation):
\begin{equation}
    \label{eq:elbo_base_skill}
    \begin{split}
    \mathcal{L}_{\textrm{ELBO}} & = \overbrace{\mathbb{E}_{z \sim q_{\boldsymbol{\phi}}(z|x)}[\log p_{\boldsymbol{\theta}}(x|z)]}^{\textrm{reconstruction loss}} - \beta_1 \overbrace{D_{KL} (q_{\boldsymbol{\phi}}(z|x)||p(z))}^{\textrm{regularizer ($\mathcal{L}_{\textrm{reg.}}$)}}\\
    & - \beta_2 \sum_{k} q(y=k|x) \underbrace{D_{KL}(q_{\boldsymbol{\phi}}(z|x)||p_{\boldsymbol{\kappa}}(z|y=k))}_{\textrm{base-skill regularizer ($\mathcal{L}_\textrm{skill}$)}}
    \end{split}\text{,}
\end{equation}
where $p_{\boldsymbol{\theta}}(x|y,z)$ and $q_{\boldsymbol{\phi}}(z|x)$ are the decoder and encoder networks with parameters $\boldsymbol{\theta}$ and $\boldsymbol{\phi}$, respectively. We also define a network $p_{\boldsymbol{\kappa}}(z|y)$ with parameters $\boldsymbol{\kappa}$ for locating the base skills in the latent skill space. $q(y=k|x)$ is calculated by Equation \eqref{eq:base skill label}. The hyperparameters $\beta_1$ and $\beta_2$ are introduced to weigh the regularizer terms. 
$\mathcal{L}_{\textrm{ELBO}}$ can be interpreted as follows. On the one hand, we intend to achieve higher reconstruction accuracy. As the reconstruction improves, our approximated posterior will also become more accurate. On the other hand, the two introduced regularizers contribute to a more structured latent skill space. The first regularizer,  $D_{KL} (q_{\boldsymbol{\phi}}(z|x)||p(z))$,  constrains the encoded distribution to be close to the prior distribution $p(z)$. Likewise, the second regularizer,  $D_{KL}(q_{\boldsymbol{\phi}}(z|x)||p_{\boldsymbol{\kappa}}(z|y))$, draws the encoded distribution nearer to the prior distribution of its corresponding base skill class. 

The learning procedure is illustrated in Figure \ref{fig:VAE_Skill_Priors} and the overall algorithm can be found in Algorithm \ref{alg:CSE-SP}. After training, we obtain a skill generator $f_{\boldsymbol{\theta}} = p_{\boldsymbol{\theta}}(x|z)$, which maps the skill embedding to the corresponding action sequence. Since there exists such a one-to-one mapping relationship, the action space $\mathcal{A}_\textrm{skill}$ is equivalent to $\mathcal{A}_z \in \mathbb{R}^{N_z}$, where $N_z$ is the skill embedding dimension. The agent should select one skill embedding in the latent space at each timestep rather than one action sequence that we typically consider. Additionally, we have the base skill locator $f_{\boldsymbol{\kappa}} = p_{\boldsymbol{\kappa}}(z|y)$ to identify the position of base skill distributions within the skill latent space. Their parameters are frozen during the later imitation learning process. 
\begin{figure}[t]
\begin{center}
    \begin{adjustbox}{width=0.4\textwidth}
    \begin{tikzpicture}
        \begin{axis}[
            title=Skill Latent Space,
            grid=major,
            width=10cm,
            height=8cm
        ]
        
        \addplot[
            only marks,
            mark=*,
            light-blue,
            opacity=0.7
        ]
        table[x=x, y=y, col sep=comma] {data/translation.csv};
        \addlegendentry{Translation}
        \addplot[
            only marks,
            mark=*,
            light-green,
            opacity=0.7
        ]
        table[x=x, y=y, col sep=comma] {data/rotation.csv};
        \addlegendentry{Rotation}
        \addplot[
            only marks,
            mark=*,
            light-red,
            opacity=0.7
        ]
        table[x=x, y=y, col sep=comma] {data/grasping.csv};
        \addlegendentry{Grasping}

        \end{axis}
    \end{tikzpicture}
    \end{adjustbox}
    \caption{\small t-SNE visualization of skill latent space.}
    \label{fig:skill_latent_space}
    \vspace{-1.5em}
\end{center}
\end{figure}

A visualization of the skill latent space helps with understanding. An illustration of the skill latent space by performing the t-SNE algorithm can be found in Figure \ref{fig:skill_latent_space}. As the figure demonstrates, three clusters are labeled with different colors, indicating three base skills we define. Each point indicates a skill embedding $z \in \mathcal{A}_z$ that corresponds to an action sequence with the length of $N_h$. A single skill embedding could encompass various base skill features, given that the skill's latent space is continuous. Consequently, a skill embedding between two base skill clusters would encompass features from both of these base skills.

\begin{algorithm}[htbp]
\caption{Learning Continuous Skill Embeddings with Base Skill Priors}\label{alg:CSE-SP}
\begin{algorithmic}[1]

\STATE Given: 
\begin{itemize}
    \item $\mathcal{D}: \{(a_0,a_1,...,a_{H-1})\}$: A Play dataset full of action sequences with horizon $H$.
    \item $\mathcal{F} = \{f_{\boldsymbol{\phi}}, f_{\boldsymbol{\theta}}, f_{\boldsymbol{\kappa}}\}$. They are the encoder network with parameters $\boldsymbol{\phi}$, the decoder network, also denoted as skill generator network with parameters $\boldsymbol{\theta}$, and the base skill locator network with parameters $\boldsymbol{\kappa}$. 
\end{itemize} 
\STATE Randomly initialize model parameters $\{\boldsymbol{\theta}, \boldsymbol{\phi}, \boldsymbol{\kappa} \}$

\WHILE{not done}
    \STATE Sample an action sequence $ x \sim \mathcal{D}$
    \STATE Encode this sequence with $f_{\boldsymbol{\phi}} = q_{\boldsymbol{\phi}}(z|x)$
    \STATE Compute the base skill distributions $f_{\boldsymbol{\kappa}} = p_{\boldsymbol{\kappa}}(z|y)$.
    \STATE Sample one latent embedding $z \sim q_{\boldsymbol{\phi}}(z|x)$
    \STATE Feed the sampled $z$ into the decoder $f_{\boldsymbol{\theta}} = p_{\boldsymbol{\theta}}(x|z)$ to get the reconstructed action sequence $\hat{x}$
    \STATE Compute the loss based on Equation \eqref{eq:elbo_base_skill}
    \STATE Update parameters $\boldsymbol{\theta}, \boldsymbol{\phi}, \boldsymbol{\kappa}$ to minimize $\mathcal{L}$
\ENDWHILE
\end{algorithmic}
\end{algorithm}
\vspace{-1em}
\subsection{Imitation Learning with Base Skill Priors}
\tikzstyle{parallelogram} = [trapezium, trapezium left angle=120, trapezium right angle=60]
\tikzstyle{hulc-encoder} = [rectangle, rounded corners, text centered, draw=black, fill=network-blue, minimum height=8cm]

\tikzstyle{skill-selecter} = [rectangle, rounded corners, text centered, draw=black, fill=network-blue, minimum height=2.75cm]

\tikzstyle{skill-labeler} = [rectangle, rounded corners, text centered, draw=black, fill=network-blue, minimum height=2.75cm]

\tikzstyle{skill-locator} = [rectangle, rounded corners, text centered, draw=black, fill=network-blue!50, minimum height=1.4cm]

\tikzstyle{skill-generator} = [rectangle, rounded corners, text centered, draw=black, fill=network-blue!50, minimum height=2cm, minimum width=4cm]

\tikzstyle{language} = [rectangle, rounded corners, text width=2.5cm, minimum height=1cm, draw=black, text centered, fill=white]
\tikzstyle{language-real} = [rectangle, rounded corners, text width=2.3cm, minimum height=1cm, draw=black, text centered, fill=white]

\tikzstyle{plan-emb} = [rectangle, minimum width=0.5cm, minimum height=1cm, draw=black, fill=plan-emb-c]
\tikzstyle{language-goal-emb} = [rectangle, minimum width=0.5cm, minimum height=1cm, draw=black, fill=language-goal-emb-c]
\tikzstyle{static-obs-emb} = [rectangle, minimum width=0.5cm, minimum height=1.cm, draw=black, fill=static-obs-emb-c]
\tikzstyle{gripper-obs-emb} = [rectangle, minimum width=0.5cm, minimum height=1cm, draw=black, fill=gripper-obs-emb-c]
\tikzstyle{language-emb} = [rectangle, minimum width=0.5cm, minimum height=1cm, draw=black, fill=language-emb-c]
\tikzstyle{dash-box1} = [rectangle, dashed, thick, minimum width=0.8cm, minimum height=6.5cm, draw=black]
\tikzstyle{dash-box2} = [rectangle, dashed, thick, minimum width=1.1cm, minimum height=4.75cm, draw=black]
\tikzstyle{dash-box3} = [rectangle, dashed, thick, minimum width=4cm, minimum height=2cm, draw=light-blue]
\tikzstyle{dash-box4} = [rectangle, dashed, thick, minimum width=9cm, minimum height=3.5cm, draw=black]
\tikzstyle{skill-emb} = [rectangle, minimum width=0.5cm, minimum height=1.25cm, draw=black,inner sep=0.5mm, fill=white]
\tikzstyle{action} = [rectangle, minimum width=0.7cm, minimum height=0.5cm, draw=black,inner sep=0.2mm, fill=white]
\tikzstyle{regularizer} = [rectangle, minimum width=1.5cm, minimum height=1.24cm, draw=black,inner sep=0.5mm, fill=white]
\tikzstyle{arrow} = [thick,->,>=stealth]

\begin{figure*}[ht]
    \centering
        \begin{adjustbox}{width=1.0\textwidth}
            \begin{tikzpicture}[node distance=2cm]
	%%%%%%%%%%%%%%%%%%%%%%%%%%%%%%%%%%%%%%%%%%%%%%% Encoder %%%%%%%%%%%%%%%%%%%%%%%%%%%%%%%%%%%%%%%%%%%%%%%
                \node (encoder-module) [rectangle, rounded corners, fill=light-yellow!50, minimum height=10.5cm, minimum width=8cm, label={[label distance=0cm]90:\Large Encoder}] {};
                
                \node[inner sep=0pt, xshift=-2.25cm, yshift=3.25cm ] (static) at (0,0)
                {\includegraphics[width=2.5cm]{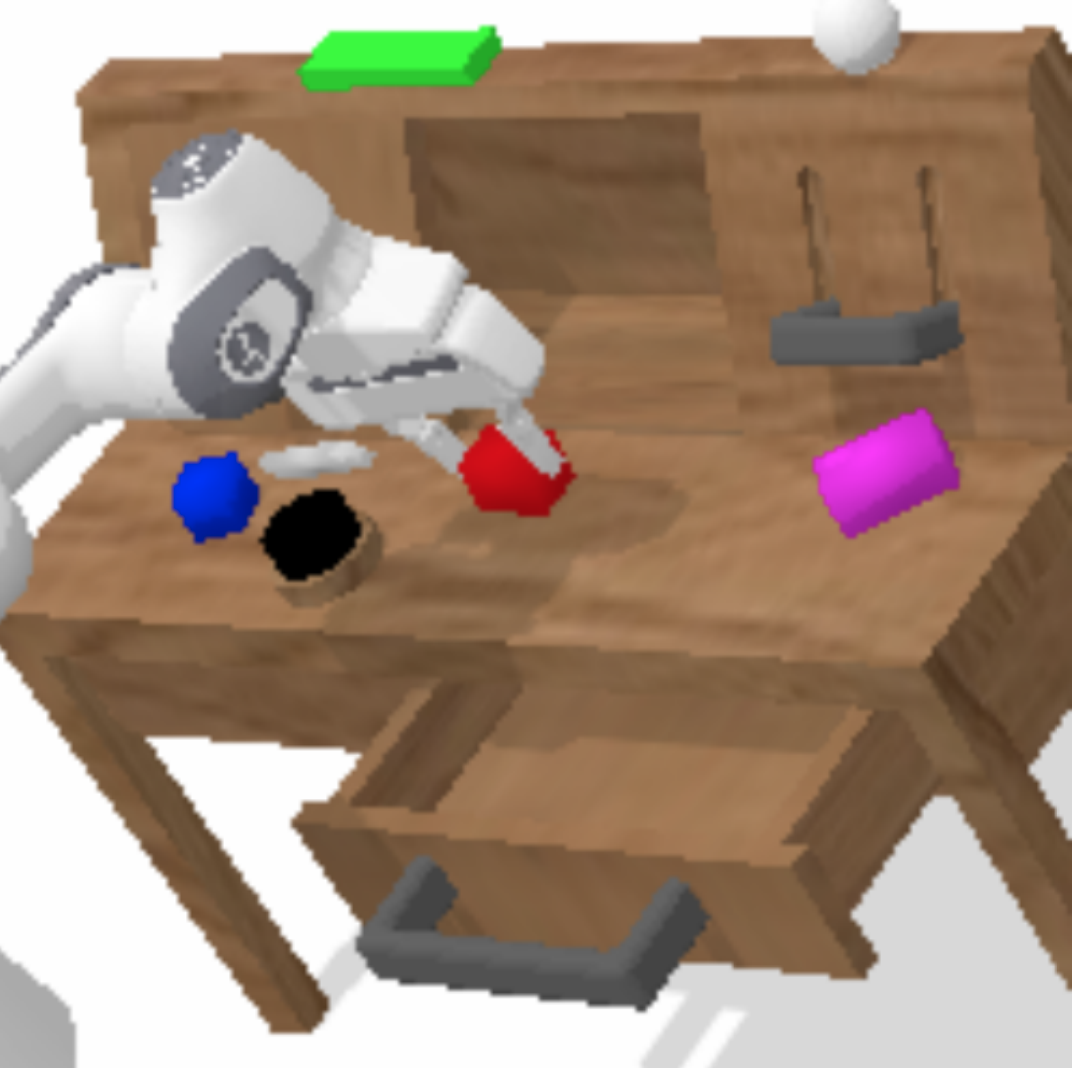}};
                
                \node[inner sep=0pt, below of=static, yshift=-1.5cm] (gripper)
                {\includegraphics[width=2.5cm]{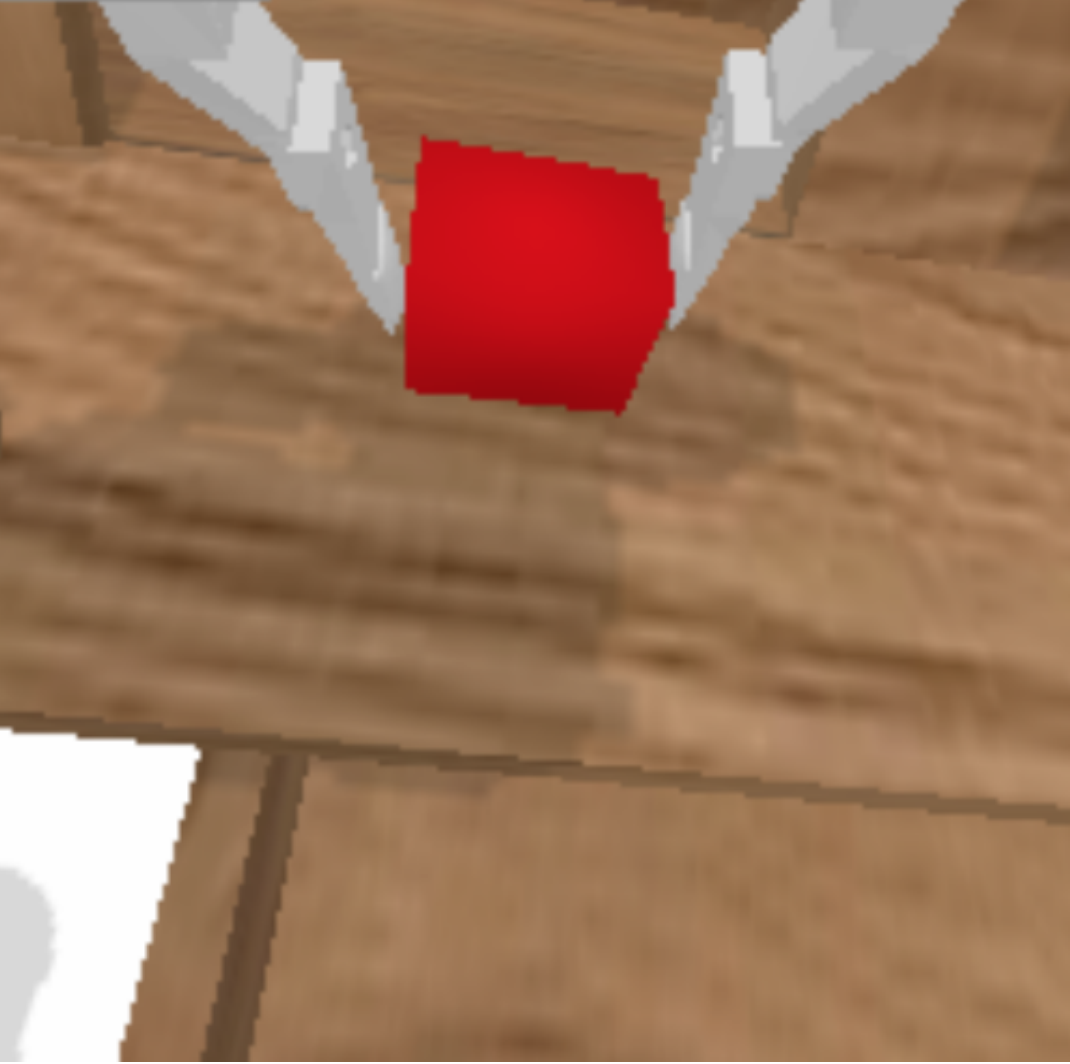}};
                
                \node (language-instruction) [language, below of=gripper, yshift=-0.75cm]  {\large Language Instruction};
                 
                \node (legend1) [rectangle, below of=language-instruction, fill=plan-emb-c, yshift=1cm, xshift=-1.25cm, draw=black, label={[label distance=0cm]0: plan emb.}]  {};
                \node (legend2) [rectangle, right of=legend1, fill=language-goal-emb-c, draw=black, xshift=0cm, label={[label distance=0cm]0: lang. goal emb.}]  {};
                \node (legend3) [rectangle, right of=legend2, fill=static-obs-emb-c, draw=black, xshift=0.75cm, label={[label distance=0cm]0: static obs. emb.}]  {};
                \node (legend4) [rectangle, below of=legend1, fill=gripper-obs-emb-c, draw=black, yshift=1.5cm, label={[label distance=0cm]0: gripper obs. emb.}]  {};
                \node (legend4) [rectangle, right of=legend4, fill=language-emb-c, draw=black, xshift=1.25cm, label={[label distance=0cm]0: language emb.}]  {};
                
                \node (hulc-encoder) [hulc-encoder, right of=gripper, xshift=1.2cm, yshift=0.75cm, text width=1.5cm, align=center] {\large Encoder Module};
                \node (f-Phi) [above left of=hulc-encoder, yshift=2.3cm, xshift=0.8cm]{$f_{\boldsymbol{\Phi}}$};
                \node (plan-emb) [plan-emb, right of=hulc-encoder, yshift=3.5cm]{};
                \node (language-goal-emb) [language-goal-emb, right of=hulc-encoder, yshift=1.75cm]{};
                \node (static-obs-emb) [static-obs-emb, right of=hulc-encoder, yshift=0cm]{};
                \node (gripper-obs-emb) [gripper-obs-emb, right of=hulc-encoder, yshift=-1.75cm]{};
                \node (language-emb) [language-emb, right of=hulc-encoder, yshift=-3.5cm]{};
                \node (dash-box1) [dash-box1, right of=hulc-encoder, yshift=0.875cm]{};
                \node (dash-box2) [dash-box2, right of=hulc-encoder, yshift=-1.75cm]{};
    %%%%%%%%%%%%%%%%%%%%%%%%%%%%%%%%%%%%%%%%%%%%%%%%%%%%%%%%%%%%%%%%%%%%%%%%%%%%%%%%%%%%%%%%%%%%%%%%%%%%%%%%
                 
    %%%%%%%%%%%%%%%%%%%%%%%%%%%%%%%%%%% SKILL-Based Imitation Learning %%%%%%%%%%%%%%%%%%%%%%%%%%%%%%%%%%%%%
                \node (skill-based-imitation-learning-module) [rectangle, rounded corners, fill=light-green!20, minimum height=10.5cm, minimum width=10cm, right of=encoder-module, xshift=7.5cm, label={[label distance=0cm]90:\Large Imitation Learning with Skill Priors}] {};
                
                \node (skill-selecter) [skill-selecter,  right of=encoder-module, text width=1.55cm, align=center, xshift=4cm, yshift=3.25cm] {Skill Emb. \large Selector};
                \node (f-lambda) [above left of=skill-selecter, yshift=-0.3cm, xshift=0.8cm]{$f_{\boldsymbol{\lambda}}$};
                \node (skill-locator) [skill-locator,  right of=encoder-module, text width=1.5cm, align=center, xshift=4cm, yshift=0.5cm] { Base Skill \large Locator};
                \node (f-kappa) [above left of=skill-locator, yshift=-0.45cm, xshift=0.8cm]{$f_{\boldsymbol{\kappa}}$};
                \node (snowflake1) [above right of=skill-locator, yshift=-0.9cm, xshift=-0.75cm]{\includegraphics[width=0.3cm]{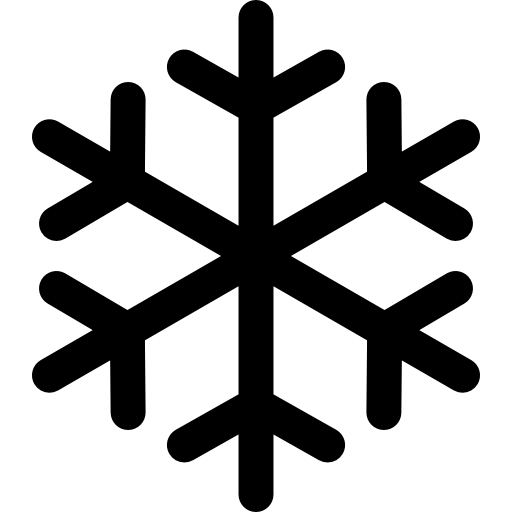}};
                \node (skill-labeler) [skill-labeler, right of=encoder-module, text width=1.5cm, align=center, xshift=4cm, yshift=-2.25cm] {Base Skill \large Selector};
                \node (f-omega) [above left of=skill-labeler, yshift=-0.3cm, xshift=0.8cm]{$f_{\boldsymbol{\omega}}$};
                \node (skill-emb1) [skill-emb, right of=skill-selecter, yshift=0.1cm, xshift=0.2cm] {$z_1$};
                \node (skill-emb2) [skill-emb, right of=skill-selecter, yshift=0cm, xshift=0.3cm] {$z_2$};
                 \node (skill-emb3) [skill-emb, right of=skill-selecter, yshift=-0.1cm, xshift=0.4cm, fill=emb-c] {$z_i$};
                 
                \node (label-emb1) [skill-emb, right of=skill-labeler, yshift=0.1cm, xshift=0.2cm] {$y_1$};
                \node (label-emb2) [skill-emb, right of=skill-labeler, yshift=0cm, xshift=0.3cm] {$y_2$};
                \node (label-emb3) [skill-emb, right of=skill-labeler, yshift=-0.1cm, xshift=0.4cm, fill=emb-c] {$y_i$};
                \node (regularizer) [regularizer, below of=skill-emb2, yshift=-0.75cm] {\Large $\mathcal{L}_{skill}$};
                \node (skill-generator) [skill-generator, right of=skill-emb3, align=center, yshift=0.2cm, xshift=1.5cm] {\large Skill Generator};
                \node (f-theta) [above left of=skill-generator, yshift=-0.65cm, xshift=-0.35cm]{$f_{\boldsymbol{\theta}}$};
                \node (snowflake2) [ above right of=skill-generator, yshift=-0.75cm, xshift=0.25cm] {\includegraphics[width=0.5cm]{figures/snowflake.png}};
                \node (action11) [action, below of=skill-generator, yshift=-0.75cm, xshift=-1.5cm] {};
                \node (action12) [action, below of=skill-generator, yshift=-0.85cm, xshift=-1.4cm] {};
                \node (action13) [action, below of=skill-generator, yshift=-0.95cm, xshift=-1.3cm, fill=emb-c] {$a_0$};
                 
                \node (action21) [action, below of=skill-generator, yshift=-0.75cm, xshift=-0.3cm] {};
                \node (action22) [action, below of=skill-generator, yshift=-0.85cm, xshift=-0.2cm, label={[label distance=0.1cm]0:\large ...}] {};
                \node (action23) [action, below of=skill-generator, yshift=-0.95cm, xshift=-0.1cm, fill=emb-c] {$a_H$};
                 
                \node (action31) [action, below of=skill-generator, yshift=-0.75cm, xshift=1.1cm] {};
                \node (action32) [action, below of=skill-generator, yshift=-0.85cm, xshift=1.2cm] {};
                \node (action33) [action, below of=skill-generator, yshift=-0.95cm, xshift=1.3cm, fill=emb-c] {$a_{nH}$};
                \node (dash-box3) [dash-box3, below of=skill-generator, yshift=-0.85cm]{};

                \node (regularizer2) [regularizer, right of=label-emb2, xshift=0.5cm] {\Large $\mathcal{L}_{cat.}$};
                \node (reconstruction) [regularizer, right of=regularizer2, xshift=0.3cm] {\Large $\mathcal{L}_{huber}$};
                
                \node (gaction1) [action, below of=reconstruction, yshift=0.2cm, xshift=-0.2cm] {};
                \node (gaction2) [action, below of=reconstruction, yshift=0.1cm, xshift=-0.1cm] {};
                \node (gaction3) [action, below of=reconstruction, yshift=0.0cm, xshift=0.0cm] {};
                \node (gaction4) [action, below of=reconstruction, yshift=-0.1cm, xshift=0.1cm] {};
                \node (gaction5) [action, below of=reconstruction, yshift=-0.2cm, xshift=0.2cm, fill=gt-c!50] {$a_{t}$};
                 
                \node (legend6) [rectangle, below of=skill-labeler, fill=emb-c, yshift=-0.5cm, xshift=-0.75cm, draw=black, label={[label distance=0cm]0: embeddings}]  {};
                \node (legend7) [rectangle, right of=legend6, fill=gt-c!60, xshift=0.75cm, draw=black, label={[label distance=0cm]0: ground truth}]  {};
    %%%%%%%%%%%%%%%%%%%%%%%%%%%%%%%%%%%%%%%%%%%%%%%%%%%%%%%%%%%%%%%%%%%%%%%%%%%%%%%%%%%%%%%%%%%%%%%%%%%%%%%%
                
    %%%%%%%%%%%%%%%%%%%%%%%%%%%%%%%%%%%%%%%%%% Real World Experiment %%%%%%%%%%%%%%%%%%%%%%%%%%%%%%%%%%%%%%%
                \node (real-world-experiment) [rectangle, rounded corners, fill=light-red!10, minimum height=10.5cm, minimum width=10cm, right of=skill-based-imitation-learning-module, xshift=8.5cm, label={[label distance=0cm]90:\Large Real-world Experiment}] {};
                \node[inner sep=0pt, below of=real-world-experiment,yshift=-0.25cm] (real-agent)
                {\includegraphics[width=9cm]{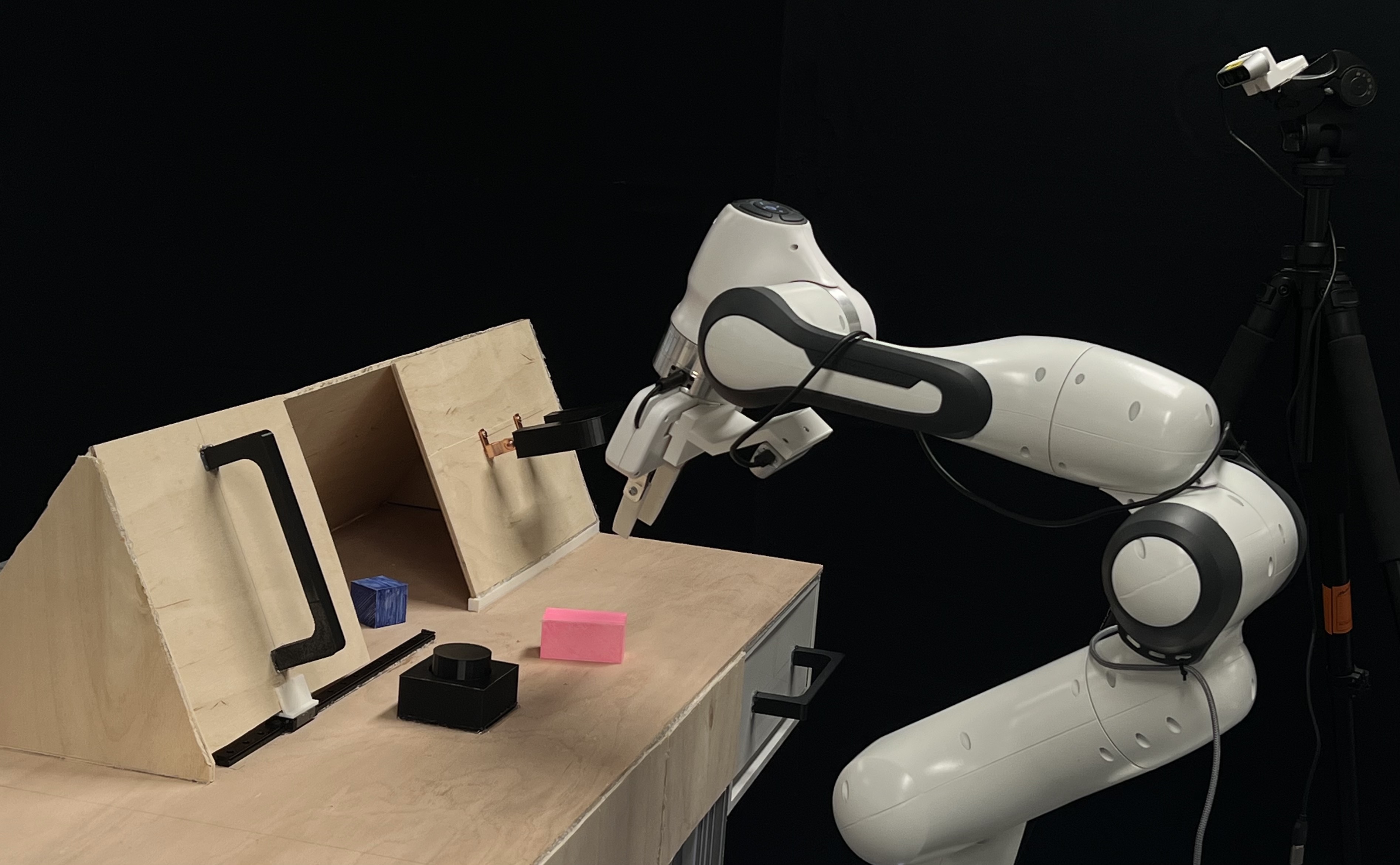}};
                
                \node[inner sep=0pt, above of=real-world-experiment, xshift=2.7cm, yshift=0.75cm] (static-real)
                {\includegraphics[width=2.5cm]{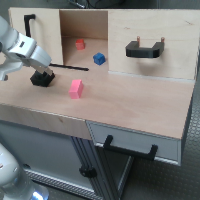}};
                
                \node[inner sep=0pt, above of=real-world-experiment, yshift=0.75cm] (gripper-real)
                {\includegraphics[width=2.5cm]{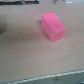}};
                
                \node (language-instruction-real) [language-real, above of=real-world-experiment, xshift=-2.7cm, yshift=0.75cm]  {\large Language Instruction};
                \node(robot-circle) [circle, minimum size=0.1cm,fill=light-blue, opacity=0.9, above of=real-agent, yshift=-1cm,xshift=0.1cm, label={[label distance=0cm, text=light-blue]135: Agent}]{};
                \node(gripper-camera-circle) [circle, minimum size=0.1cm,fill=light-red, opacity=0.7, above of=real-agent, yshift=-2cm, xshift=0.6cm, label={[label distance=0cm, text=light-red, text width=2cm]-45: Gripper\\camera}]{};
                \node(static-camera-circle) [circle, minimum size=0.1cm,fill=light-red, opacity=0.7, above right of=real-agent, yshift=1cm, xshift=2.4cm, label={[label distance=0cm, text=light-red, text width=1.5cm]-90: Static\\camera}]{};
                \node(tabletop-circle) [circle, minimum size=0.1cm,fill=light-red, opacity=0.7, left of=real-agent, yshift=-0.5cm, xshift=0cm]{};
                \node(tabletop-label)[rectangle, text=light-red, text width=2cm, above of=tabletop-circle,yshift=0.5cm]{Tabletop\\environment};
                \node (dash-box4) [dash-box4, above of=real-world-experiment, yshift=0.75cm]{};

    %%%%%%%%%%%%%%%%%%%%%%%%%%%%%%%%%%%%%%%%%%%%%%%%%%%%%%%%%%%%%%%%%%%%%%%%%%%%%%%%%%%%%%%%%%%%%%%%%%%%%%%%
                
                \draw [arrow] (static.east) -- (static.east-|hulc-encoder.west);
                \draw [arrow] (gripper.east) -- (gripper.east-|hulc-encoder.west);
                \draw [arrow] (language-instruction.east) -- (language-instruction.east-|hulc-encoder.west);
                \draw [arrow] (plan-emb.west-|hulc-encoder.east) -- (plan-emb.west);
                \draw [arrow] (language-goal-emb.west-|hulc-encoder.east) -- (language-goal-emb.west);
                \draw [arrow] (static-obs-emb.west-|hulc-encoder.east) -- (static-obs-emb.west);
                \draw [arrow] (gripper-obs-emb.west-|hulc-encoder.east) -- (gripper-obs-emb.west);
                \draw [arrow] (language-emb.west-|hulc-encoder.east) -- (language-emb.west);
                \draw [arrow, dashed] (skill-selecter.west-|dash-box1.east) -- (skill-selecter.west);
                \draw [arrow, dashed] (skill-labeler.west-|dash-box2.east) -- (skill-labeler.west);
                \draw [arrow] (skill-emb2.west-|skill-selecter.east) -- (skill-emb2.west);
                \draw [arrow] (regularizer.west-|skill-locator.east) -- (regularizer.west);
                \draw [arrow] (label-emb2.west-|skill-labeler.east) -- (label-emb2.west);
                \draw [decorate,decoration={brace, amplitude=1mm, raise=.5mm}] (label-emb1.north east) -- (label-emb3.north east) node[above right, midway] {\scriptsize $n+1$};
                \draw [decorate,decoration={brace,amplitude=1mm, raise=.5mm}] (skill-emb1.north east) -- (skill-emb3.north east) node[above right, midway] {\scriptsize $n+1$};
                \begin{scope}[transform canvas={yshift=-0.15cm}]
                    \draw [arrow, dashed] (skill-emb3.east) -- (skill-emb3.east-|skill-generator.west);
                \end{scope}
                \begin{scope}[transform canvas={yshift=0cm}]
                    \draw [arrow, dashed] (skill-emb3.east) -- (skill-emb3.east-|skill-generator.west) node[above,midway] {...};
                \end{scope}
                \begin{scope}[transform canvas={yshift=.3cm}]
                    \draw [arrow, dashed] (skill-emb3.east) -- (skill-emb3.east-|skill-generator.west);
                \end{scope}
                \draw [arrow, dashed] (skill-generator.south-|action13.north) -- (action13.north);
                \draw [arrow, dashed] (skill-generator.south-|action23.north) -- (action23.north);
                \draw [arrow, dashed] (skill-generator.south-|action33.north) -- (action33.north);
                \draw [arrow] (action11.north) -- +(-0.15,0.3) -- +(0.4, 0.3) -- +(1.0, -1) -- +(1.55, -1) -- (action23.south);
                \draw [arrow] (action21.north) -- +(-0.15,0.3) --  +(0.45,0.3)  -- + (0.65, -0.1) + (0.9, -0.6) -- + (1.1, -1) -- +(1.75, -1) -- (action33.south);
                \draw [decorate,decoration={brace, amplitude=1mm, raise=.5mm}] (action31.north east) -- (action33.north east) node[above right, midway] {\scriptsize $H$};

                \draw [arrow] (skill-emb3.south-| regularizer.north) -- (regularizer.north);
                \draw [arrow] (label-emb2.north-| regularizer.south) -- (regularizer.south);
                \draw [arrow] (label-emb3.east|- regularizer2.west) -- (regularizer2.west);
                \draw [arrow, dashed, draw=light-blue] (dash-box3.south-| reconstruction.north) -- (reconstruction.north);
                \draw [arrow] (gaction3.north-| reconstruction.south) -- (reconstruction.south);
                \draw [arrow, dashed, draw=light-blue] (dash-box3.east) -- +(+1.25, 0.0) |- (robot-circle.west);
                \draw [arrow, draw=light-red] (gripper-camera-circle.north) -- +(0.0, 2.85) -| (gripper-real.south);
                \draw [arrow, draw=light-red] (static-camera-circle.north) -- +(0.0, 0.45) -| (static-real.south);
                \draw [arrow, dashed] (dash-box4.north) -- +(0.0, 0.5) -| (hulc-encoder.north);
                \draw [arrow, draw=light-red] (tabletop-label.south) -- (tabletop-circle.north);
                % \draw [arrow] (skill-generator.east) -- (gaction3.west);
	        \end{tikzpicture}
        \end{adjustbox}

        \caption{\small The Overall Architecture. Following the encoding process, the static observation, gripper observation, and language instruction are generated to embeddings for the plan, language goal, language, static observation, and gripper observation. The skill selector module subsequently decodes a sequence of skill embeddings using the plan, observation, and language goal embeddings. The skill labeler labels the skill embeddings with the base skills:  translation, rotation, and grasping. The base skill regularization loss is calculated based on the base skill prior distributions (from base skill locator $f_{\boldsymbol{\kappa}}$), selected skill instance, and labeled probability indicating its belonging to specific base skills. This labeled probability is also leveraged to determine the categorical regularization loss. Finally, the pre-trained and frozen skill generator $f_{\boldsymbol{\theta}}$ decodes all the skill embeddings into action sequences, which are then utilized to calculate the reconstruction loss (Huber loss).}
        \label{fig:spil}
    \end{figure*}
\begin{algorithm}[ht]
\caption{Imitation Learning with Skill Priors}\label{alg:SPIL}
\begin{algorithmic}[1]

\STATE Given: 
\begin{itemize}
    \item $\mathcal{D}: \{(D^{\textrm{play}}, D^{\textrm{lang}})\}$: Play Dataset and Language Dataset
    \item $\mathcal{F} = \{f_{\boldsymbol{\Phi}}, f_{\boldsymbol{\lambda}}, f_{\boldsymbol{\kappa}}, f_{\boldsymbol{\omega}}, f_{\boldsymbol{\theta}}\}$. They are the encoder $f_{\boldsymbol{\Phi}}$, the skill embedding selecter $f_{\boldsymbol{\lambda}}$, the base skill locator $f_{\boldsymbol{\kappa}}$, the base skill selector $f_{\boldsymbol{\omega}}$, the skill generator networks $f_{\boldsymbol{\theta}}$ with parameters $\boldsymbol{\Phi}$, $\boldsymbol{\lambda}$, $\boldsymbol{\kappa}$, $\boldsymbol{\omega}$, and $\boldsymbol{\theta}$, respectively.
\end{itemize} 
\STATE Randomly initialize model parameters $\{\boldsymbol{\Phi}, \boldsymbol{\lambda}, \boldsymbol{\omega} \}$
\STATE Initialize parameters $\boldsymbol{\theta}$ and $\boldsymbol{\kappa}$ with pre-trained skill generator and base skill locator 
\STATE Freeze the parameters $\boldsymbol{\theta}$ and $\boldsymbol{\kappa}$. 

\WHILE{not done}
    \STATE $\mathcal{L} \leftarrow 0$
    \FOR{$l$ in \{\textrm{play, lang}\}}
        \STATE Sample a (demonstration, context) $(x^l, c^l) \sim D^l$
        \STATE Encode the observation, goal, and plan embeddings, using  the encoder network $f_{\boldsymbol{\Phi}}$
        \STATE \textbf{Skill Embedding Selecter}  $f_{\boldsymbol{\lambda}}$ selects the skill embedding sequence
        \STATE Determinate a sequence of base skill probabilities with \textbf{Base Skill Selector} $f_{\boldsymbol{\omega}}$.
        \STATE Determinate base skill locations in the latent space with \textbf{Base Skill Locator} $f_{\boldsymbol{\kappa}}$
        \STATE \textbf{Skill Generator} $f_{\boldsymbol{\theta}}$ maps the skill embeddings to action sequences.
        \STATE Calculate the loss function $\mathcal{L}_l$ according to \eqref{eq:L_elbo spil}
        \STATE Accumulate imitation loss $\mathcal{L} \mathrel{+}= \mathcal{L}_l$
    \ENDFOR
    \STATE update parameters $\{\boldsymbol{\Phi}, \boldsymbol{\lambda}, \boldsymbol{\omega} \}$  w.r.t $\mathcal{L}$
\ENDWHILE
\end{algorithmic}
\end{algorithm}
After acquiring the skill embedding space $\mathcal{A}_z$ and the distributions of base skill priors in such latent space, we can train a policy using imitation learning based on that. This approach results in a policy with enhanced generalization capabilities, as incorporating prior knowledge prevents the model from overfitting. 
% The prior knowledge can bias the model towards solutions that align with our prior understanding, thus preventing the model from learning irrelevant or noisy patterns in the data that may lead to overfitting. 
The base skill priors we have defined encapsulate human proficiency in task completion. We aim to leverage the prior knowledge contained in base skills to reduce the agent's reliance solely on training data. In our approach, the agent learns to choose a skill that embodies motion-related human knowledge instead of determining the action at every step. Meanwhile, it also selects the appropriate base skill for the current state, mirroring the habitual approach of humans in accomplishing tasks. 
% An more detailed example is demonstrated in Figure Take an example of 'Lifting the ' \textcolor{red}{TODO:}

We extend the idea of MCIL \cite{lynch2021language} and HULC \cite{9849097} by employing an action space $\mathcal{A}_z$ comprising skill embeddings instead of Cartesian action space $\mathcal{A}$. In this framework, the action performed by the agent is no longer a single 7 DoF movement in one time step, but instead, a skill (action sequence) over a horizon $N_h$. Consequently, the agent learns to select a skill based on the current observation. After the skill is performed, the agent selects the next skill based on the subsequent observation, and the process continues iteratively until the agent completes the task or the time runs out.
Figure \ref{fig:spil} depicts the overall structure of our approach. Given the superior performance of the HULC model, we employ its encoder, denoted as $f_{\boldsymbol{\Phi}}$, to transform the static observation, gripper observation, and language instruction into their corresponding embeddings. All these embeddings align with the definition provided in the HULC model. Additionally, to extract the overall process information from the language instruction, we introduce extra language embedding.
% It is important to note that we extract two embeddings from language instruction. The first is the language goal embedding, which lies in the same latent goal space as image goal embedding. The second is the language embedding, which is utilized to infer the appropriate base skill to be employed, given the current observation. Revisiting the concept of multi-context learning \cite{lynch2021language}, our objective is to train a single imitation learning policy conditioned on either language instruction or goal observation. Therefore, the encoded language embedding and goal image embedding should be in the same space. However, this structure presents a challenge, as the language instruction contains not only the final state information but also the process information about how to complete the tasks. 
This process information is crucial for inferring the intermediate-level compositions of base skills required for successful task completion. We further analyze four key parts in our structure:

\begin{itemize}
    \item \textbf{Skill Embedding Selector}: The skill embedding selector, denoted as $f_{\boldsymbol{\lambda}}$, selects skill embeddings in the pre-trained latent space. A bidirectional LSTM network is employed for this skill embedding selector.  
    \item \textbf{Base Skill Selector}: The base skill selector $f_{\boldsymbol{\omega}}$, also a bidirectional LSTM network, determines the base skill to which a given skill belongs. 
    \item \textbf{Base Skill Locator}: The base skill locator shares the same parameters with base skill locator $f_{\boldsymbol{\kappa}}$ in Figure \ref{fig:VAE_Skill_Priors}. It has the task of locating the base skill locations in the latent space. The input to this network is a $3 \times 3$ identity matrix, signifying the one-hot representing of three base skills. These locations are used to calculate the regularization loss.
    \item \textbf{Skill Generator}: The skill generator, denoted as $f_{\boldsymbol{\theta}} = p_{\boldsymbol{\theta}}(x|z) : \mathcal{A}_z \rightarrow \mathcal{A}_{\textrm{skill}}$, shares the same parameters with the decoder component in Figure \ref{fig:VAE_Skill_Priors}. Its parameters are frozen during the imitation learning process. Its function is to transform space from skill embedding space $\mathcal{A}_z$ to skill space $\mathcal{A}_{\textrm{skill}}$. These skills (action sequences) are combined chronologically for a longer action sequence. 
\end{itemize}

The objective of our model is to learn a policy $\pi(x|s_c,s_g)$ conditioned on the current state $s_c$ and the goal state $s_g$ and outputting $x$, a sequence of actions, namely a skill. Since we introduced the base skill concept into our model, the policy $\pi(\cdot)$ should also find the best base skill $y$ for the current observation. We have $\pi(x,y|s_c,s_g)$, where $y$ is the base skill the agent chooses based on the current state and goal state. \\
Inspired by the conditional variational autoencoder (CVAE):
\begin{equation}
    \label{eq:C-VAE}
    \log p(x|c) \ge \mathbb{E}_{q(z|x,c)}[\log p(x|z,c)] - D_{KL}(q(z|x,c) || p(z|c))
\end{equation}
where c is a symbol to describe a general condition, we would like to extend the above equation by integrating $y$ which indicates the base skill. The evidence we want to maximize then turns to $p(x,y|c)$. We employ the approximate variational posterior $q(y,z|x,c)$ to approximate the intractable true posterior $p(y,z|x,c)$ where $z$ indicates the skill embeddings in the skill latent space. We intend to find the ELBO by measuring the KL divergence between the true posterior and the posterior approximation (detailed theoretical motivation in the Appendix). We have 
\begin{equation}
    \begin{split}
        \mathcal{L} & = \overbrace{\mathbb{E}_{z \sim q_{\boldsymbol{\phi}}(x,c)}\log p_{\boldsymbol{\theta}}(x|z)}^{\textrm{Reconstruction loss ($\mathcal{L}_{\textrm{huber}}$)}} \\ 
        & - \gamma_1 \sum_{k} q_{\boldsymbol{\omega}}(y=k|c) \overbrace{D_{KL}(q_{\boldsymbol{\Phi}, \boldsymbol{\lambda}}(z|x,c)||p_{\boldsymbol{\kappa}}(z|y))}^{\textrm{Base skill regularizer ($\mathcal{L}_{\textrm{skill}}$)}}\\
        & - \gamma_2 \overbrace{D_{KL}(q_{\boldsymbol{\omega}}(y|c)||p(y))}^{\textrm{Categorical regularizer ($\mathcal{L}_{\textrm{cat.}}$)}}
    \end{split}
    \label{eq:L_elbo spil}
\end{equation}
where $c$ represents a combination of the current state and the goal state $(s_c,s_g)$. $z$ is skill embedding in the latent skill space. $p_{\boldsymbol{\theta}}(x|z)$ is the skill generator network $f_{\boldsymbol{\theta}}$ with parameters $\boldsymbol{\theta}$ and it is trained by VAEs discussed in the previous session and frozen during the imitation learning. $f_{\boldsymbol{\omega}} = q_{\boldsymbol{\omega}}(y|c)$ corresponds to the skill labeller with parameter $\boldsymbol{\omega}$. $q_{\boldsymbol{\Phi}, \boldsymbol{\lambda}}(z|x,c)$ refers to the encoder network $f_{\boldsymbol{\Phi}}$ plus the skill embedding selector network $f_{\boldsymbol{\lambda}}$. Furthermore, $p_{\boldsymbol{\kappa}}(z|y)$ constitutes the base skill prior locater $f_{\boldsymbol{\kappa}}$ with parameter $\boldsymbol{\kappa}$. It is also trained by VAEs, as discussed in the previous section and frozen during the training process. Here, we use Huber loss as the metric for reconstructive loss. Intuitively, the base skill regularizer is used to regularize a skill embedding, depending on its base skill category. The categorial regularizer aims to regularize the base skill classification based on the prior categorical distribution of $y$. The overall algorithm can be seen in Algorithm \ref{alg:SPIL}.

\section{Experiments}\label{section:experiments}
\begin{table*}[ht]
    \centering
    \caption{CALVIN benchmark results}
    \label{table: CALVIN benchmark results}
    \resizebox{0.98\linewidth}{!}{
    \begin{tabular}{ c| c | c | c  c  c  c  c  c} 
        \toprule
            \multirow{3}*{Environment} & \multirow{3}*{Method} & \multirow{3}*{Train $\rightarrow$ Test}&\multicolumn{6}{c}{LH-MTLC} \\
             & & & \multicolumn{6}{c}{No. Instructions in a Row (1000 chains)} \\
        \cline{4-9}
            &  & & 1 & 2 & 3 & 4 & 5 & Avg. Len. \\
        \midrule
            \multirow{12}*{\makecell[c]{Zero-shot Multi \\ Environment}}& MCIL & A,B,C $\rightarrow$ D & 30.4\% & 1.3\% & 0.17 \% & 0\% & 0\% & 0.31 \\
            &HULC & A,B,C $\rightarrow$ D & 41.8\% (2.3)& 16.5\% (2.5)& 5.7\% (1.3)& 1.9\% (0.9)& 1.1\% (0.5)& 0.67 (0.1)\\    
            &SPIL (Ours) & A,B,C $\rightarrow$ D & \textbf{74.2}\% (1.4)& \textbf{46.3}\% (3.4)& \textbf{27.6}\% (3.4)& \textbf{14.7}\% (2.3)& \textbf{8.0}\% (1.7)& \textbf{1.71} (0.11)\\
        \cmidrule{2-9}
            & 3D Diffuser Actor*$^\dagger$ & A,B,C $\rightarrow$ D & 92.2 \% & 78.7 \% & 63.9 \% & 51.2 \% & 41.2 \% & 3.27 \\
            & GR-1*$^\dagger$  & A,B,C $\rightarrow$ D & 85.4 \% & 71.2 \% & 59.6 \% & 49.7 \% & 40.1 \% & 3.06 \\
            & SuSIE* & A,B,C $\rightarrow$ D & 87.0 \% & 69.0 \% & 49.0 \% & 38.0 \% & 26.0 \% & 2.69 \\
            & RoboFlamingo* & A,B,C $\rightarrow$ D & 82.4 \% & 61.9 \% & 46.6 \% & 33.1 \% & 23.5 \% & 2.47\\
        \cmidrule{2-9}
            &$\gamma_1=1.0 \times 10^{-2}$ & A,B,C $\rightarrow$ D & 71.3\% (1.4)& 45.8\% (3.8)& 25.4\% (2.3)& 13.1\% (0.9)& 6.5\% (0.5)& 1.62 (0.05)\\
            &$\gamma_2=1.0 \times 10^{-4}$ & A,B,C $\rightarrow$ D & 70.6\% (4.2)& 46.3\% (3.2)& 25.1\% (3.0)& 14.1\% (1.0)& 7.3\% (1.3)& 1.63 (0.08)\\   
            &$N_h=4$ & A,B,C $\rightarrow$ D & 71.4\% (2.1) & 41,0\% (3,1)& 24.1\% (1,2)& 12.1\% (1.1)& 7.4\% (0.6) & 1.58 (0.07)\\
            &$N_h=6$ & A,B,C $\rightarrow$ D & 74.0\% (1.6) & 44.2\% (2.6)& 25.2\% (2.0) & 13.0\% (2.4) & 7.9\% (1.7)& 1.65 (0.08)\\
            &w/o base skills & A,B,C $\rightarrow$ D & 57.5\% (1.8) & 27.9\% (2.1) & 12.2\% (1.2)& 5.0\% (1.1) & 2.2\% (0.7) & 1.05 (0.06)\\
        \midrule
            \multirow{3}*{\makecell[c]{Single \\ Environment}}& MCIL & D $\rightarrow$ D & 76.4\% (1.5) & 48.8\% (4.1) & 30.1\% (4.5) & 18.1\% (3.0) & 9.3\% (3.5) & 1.82 (0.2) \\
            &HULC & D $\rightarrow$ D & 82.7\% (0.3) & 64.9\% (1.7) & 50.4\% (1.5) & \textbf{38.5}\% (1.9) & 28.3\% (1.8) & 2.64 (0.05) \\
            &SPIL (Ours) & D $\rightarrow$ D & \textbf{84.6}\% (0.6) & \textbf{65.1}\% (1.3) & \textbf{50.8}\% (0.4)& 38.0\% (0.6)& \textbf{28.6}\% (0.3) & \textbf{2.67} (0.01) \\
        \bottomrule
    \end{tabular}

    }

    \raggedright \footnotesize * indicates that the model leverages pre-trained foundation models. $^\dagger$ means that the model leverages extra proprioceptive state as input
    \vspace{-1em}
\end{table*}

In this section, we present the experiments conducted to investigate the generalization ability of our model in comparison to other baselines. We choose the CALVIN \cite{mees2022calvin} benchmark to evaluate our model. 
The CALVIN benchmark is introduced to facilitate learning language-conditioned tasks across four manipulation environments: A, B, C, and D. Each environment features a Franka Emika Panda robot arm equipped with a gripper and a desk that includes a sliding door and a drawer. Additionally, the desk has a button that can toggle a green light and a switch to control a light bulb. Note that each environment has a different desk with various of textures and the position of static elements such as the sliding door, drawer, light, switch, and button are different across each environment. Experiments are conducted in two settings: (1) a single environment where the training and testing environments are the same, and (2) zero-shot multi-environments where training occurs in the first three environments and testing takes place in the fourth, previously unseen environment.

We choose long-horizon multi-task language control (LH-MTLC) to evaluate the effectiveness of the learned multi-task language-conditioned policy in accomplishing several language instructions in a row under the zero-shot multi environment. We also compare other skill-based reinforcement learning approaches to show the advantages of our approaches against theirs. 

We analyze the result of our model by comparing it to other baselines (shown in Table \ref{table: CALVIN benchmark results}). We evaluate the models with 1000 five-task chains. The columns labeled from one to five demonstrate the success rate of continuously completing that number of tasks in a row. The average length indicates the average number of tasks the agent can continuously complete when given five tasks in a row (The remaining tasks are not performed if one task fails in the middle). Subsequently, ablation studies on hyperparameters $\gamma_1$,$\gamma_2$ in \eqref{eq:L_elbo spil} and the length of skill $N_h$ (the default value is 5) are performed in the zero-shot multi-environment. Each model is evaluated three times across 3 random seeds.

% \subsection{Single Environment}
% As shown in Table \ref{table: CALVIN benchmark results (single Environment)}, our model achieves the highest score among all baselines in the Single Environment setting. Compared to the current SOTA model HULC, the success rate of completing one to five tasks in a row has increased by 1.9 \%, 0.2\%, 0.4 \%, -0.5\%, and 0.3 \% respectively. The overall average completed task length, indicating the average number of continuously completed tasks, increased from 2.64 to 2.67. It is important to note that this experiment demonstrates that in the single environment setting, our proposed method does not exhibit any performance degradation (even higher) compared to other baselines. The primary focus of our work is on the Zero-shot Multi Environment setting, which serves to demonstrate the model's generalization ability. The comparisons with other skill-based reinforcement learning approaches \cite{shiskill, pertsch2020spirl} can be found in Appendix \ref{sec:Comparison with Other Skill-based Approaches}.

\vspace{-1em}
\subsection{Environment Result}
As evidenced in Table \ref{table: CALVIN benchmark results}, our model substantially improves compared to our baselines HULC and MCIL in a zero-shot multi-environment setting. Compared to the current SOTA model HULC, the success rate of completing one to five tasks in a row has increased by 32.4\% , 29.8\%, 21.9 \%, 12.8\%, and 6.9 \%, respectively. The overall average length increased from 0.67 to 1.71. Note that zero-shot multi-environment presents a challenging environment, as the agent must solve tasks in an unfamiliar environment. The performance in this setting represents the agent's ability to truly understand and connect the concepts in language instructions with real objects and actions. The performance of our model demonstrates a significant improvement, thus confirming our hypothesis that using skill priors to learn intermediate-level task composition can improve generalization capabilities.
The other SOTA models - SuSIE \cite{black2023zeroshotroboticmanipulationpretrained}, RoboFlamingo \cite{li2023vision}, GR-1 \cite{wu2023unleashinglargescalevideogenerative}, 3D Diffuser Actor \cite{ke20243ddiffuseractorpolicy}, which leverage pre-trained foundation models, as listed in Table \ref{table: CALVIN benchmark results} for reference.
It is worth mentioning that our SPIL model also outperforms the baselines in the single environment setting.

\vspace{-1em}
\subsection{Real-world Experiments}
\begin{figure*}[ht]
\centering
\subfloat[ push button]{\includegraphics[width=.19\textwidth]{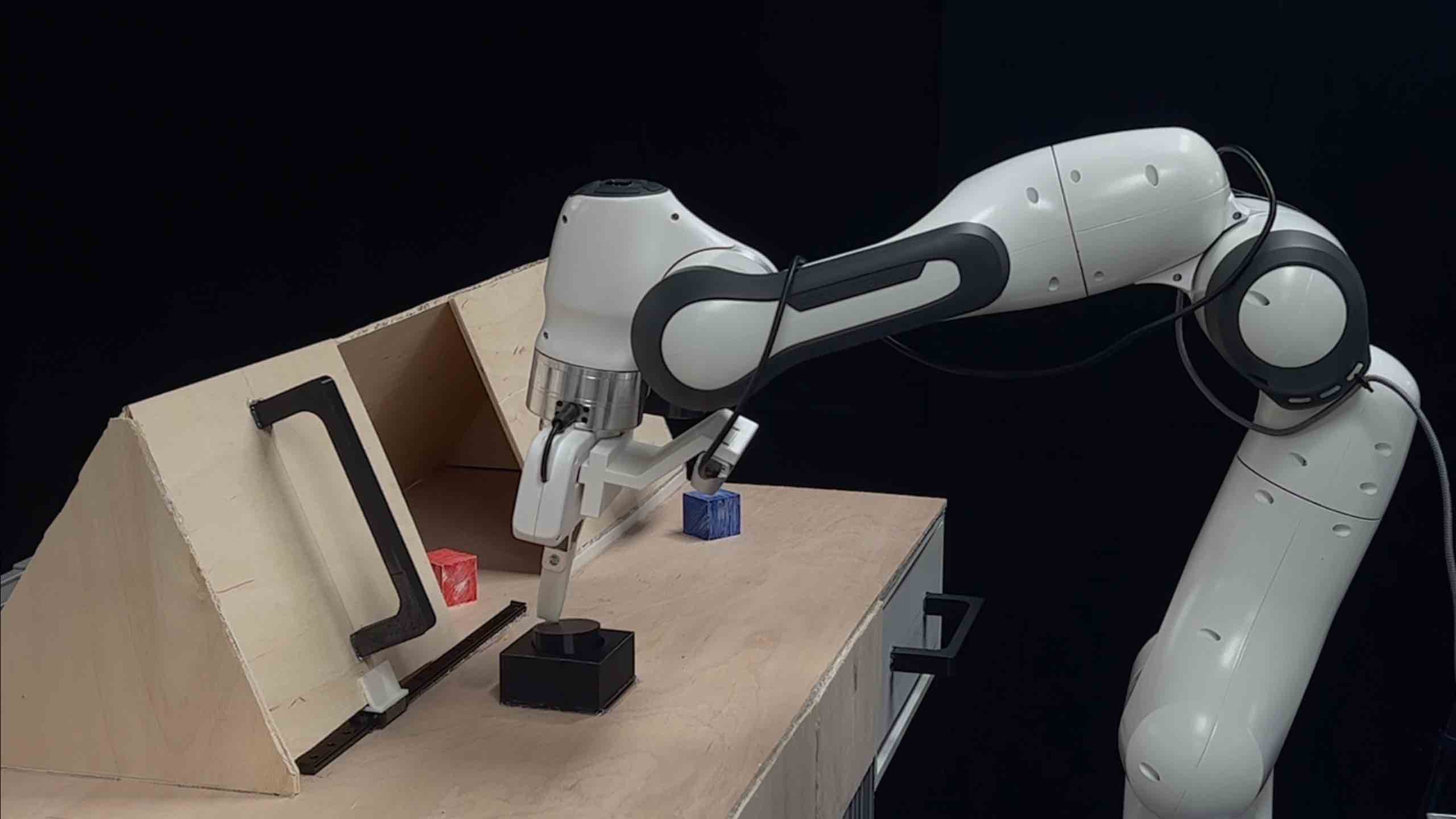}}
\hfill
\subfloat[ lift red block]{\includegraphics[width=.19\textwidth]{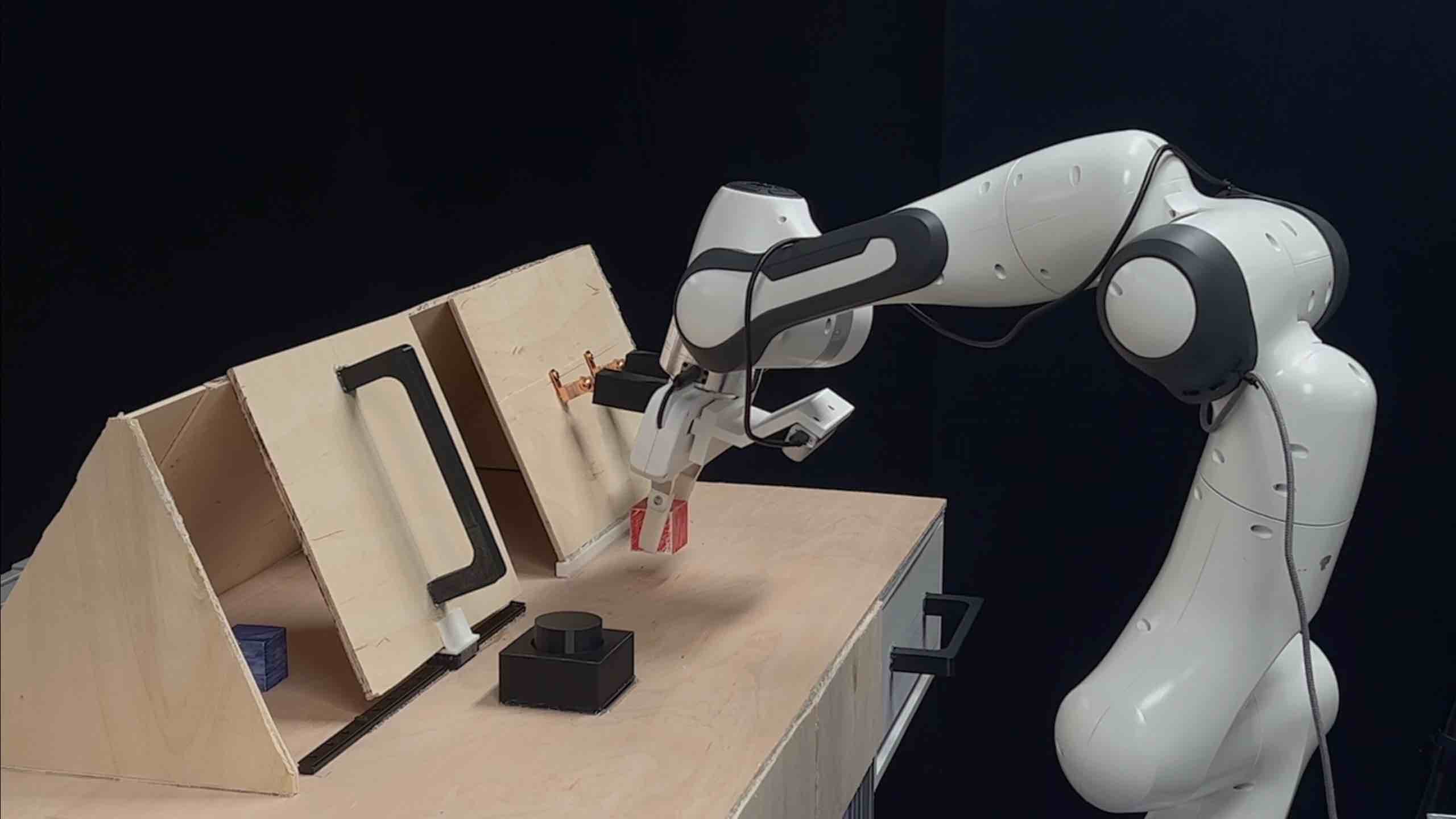}}
\hfill
\subfloat[ lift pink block]{\includegraphics[width=.19\textwidth]{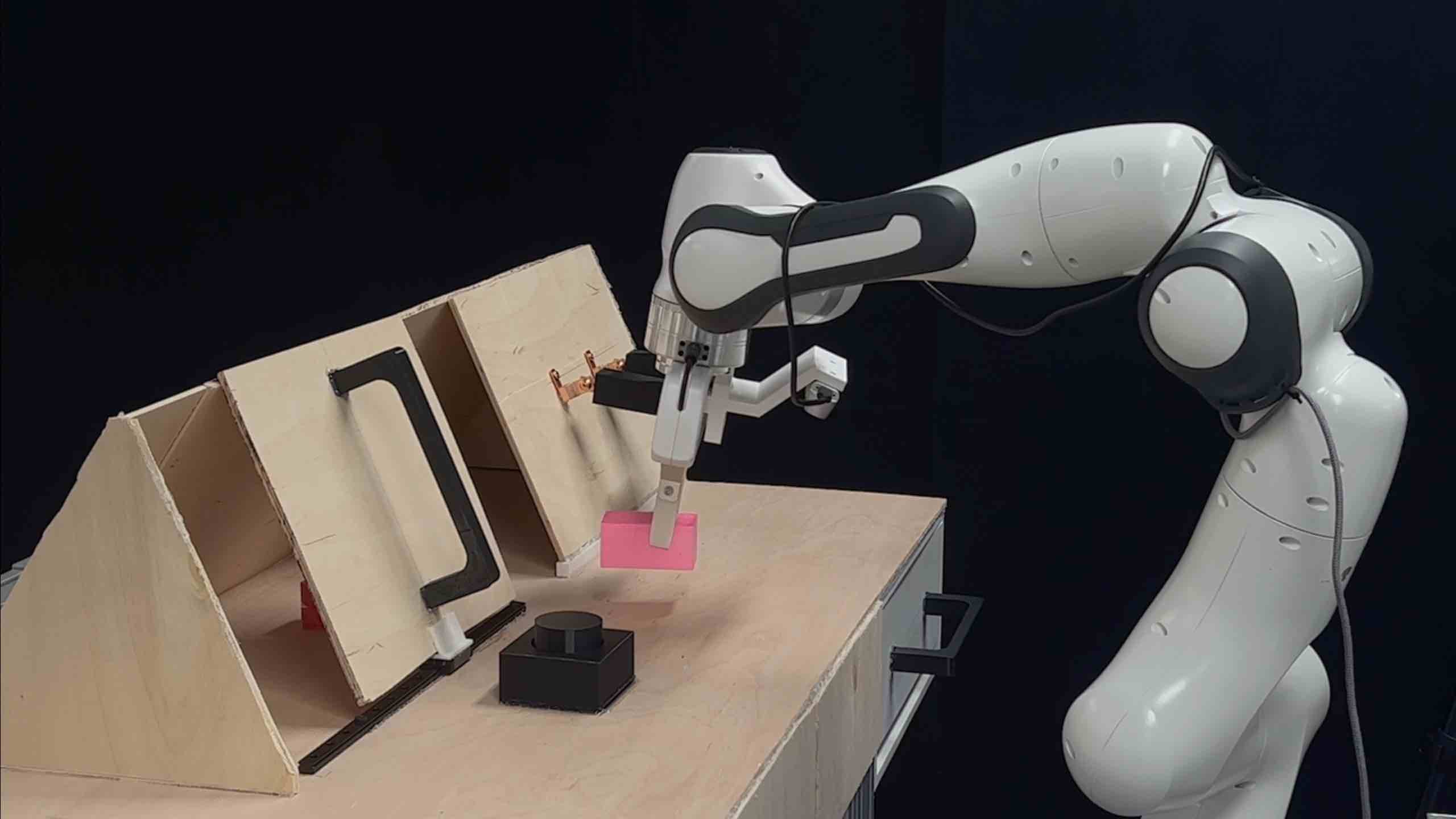}}
\hfill
\subfloat[ lift blue block]{\includegraphics[width=.19\textwidth]{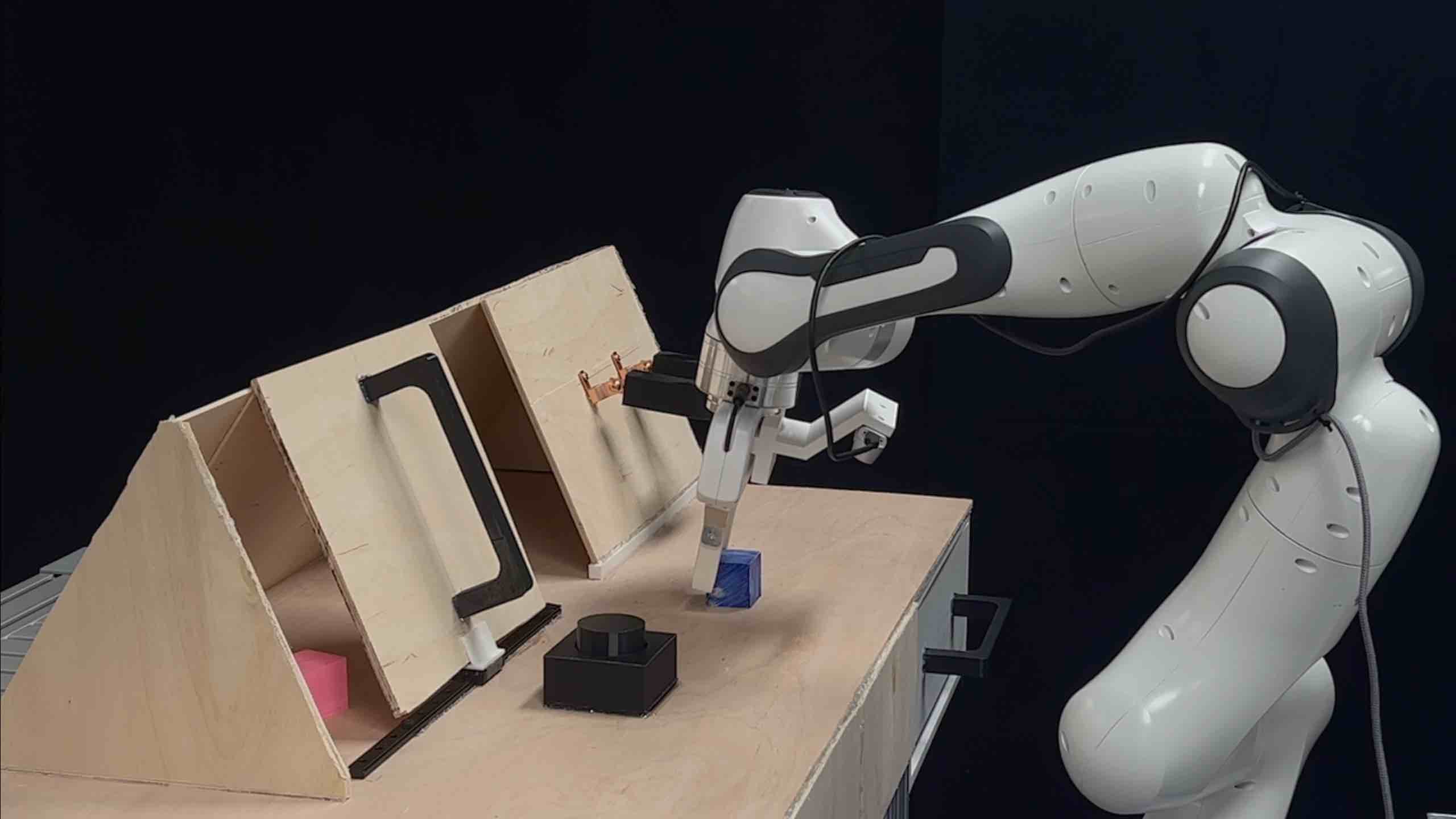}}
\hfill
\subfloat[ move slider left]{\includegraphics[width=.19\textwidth]{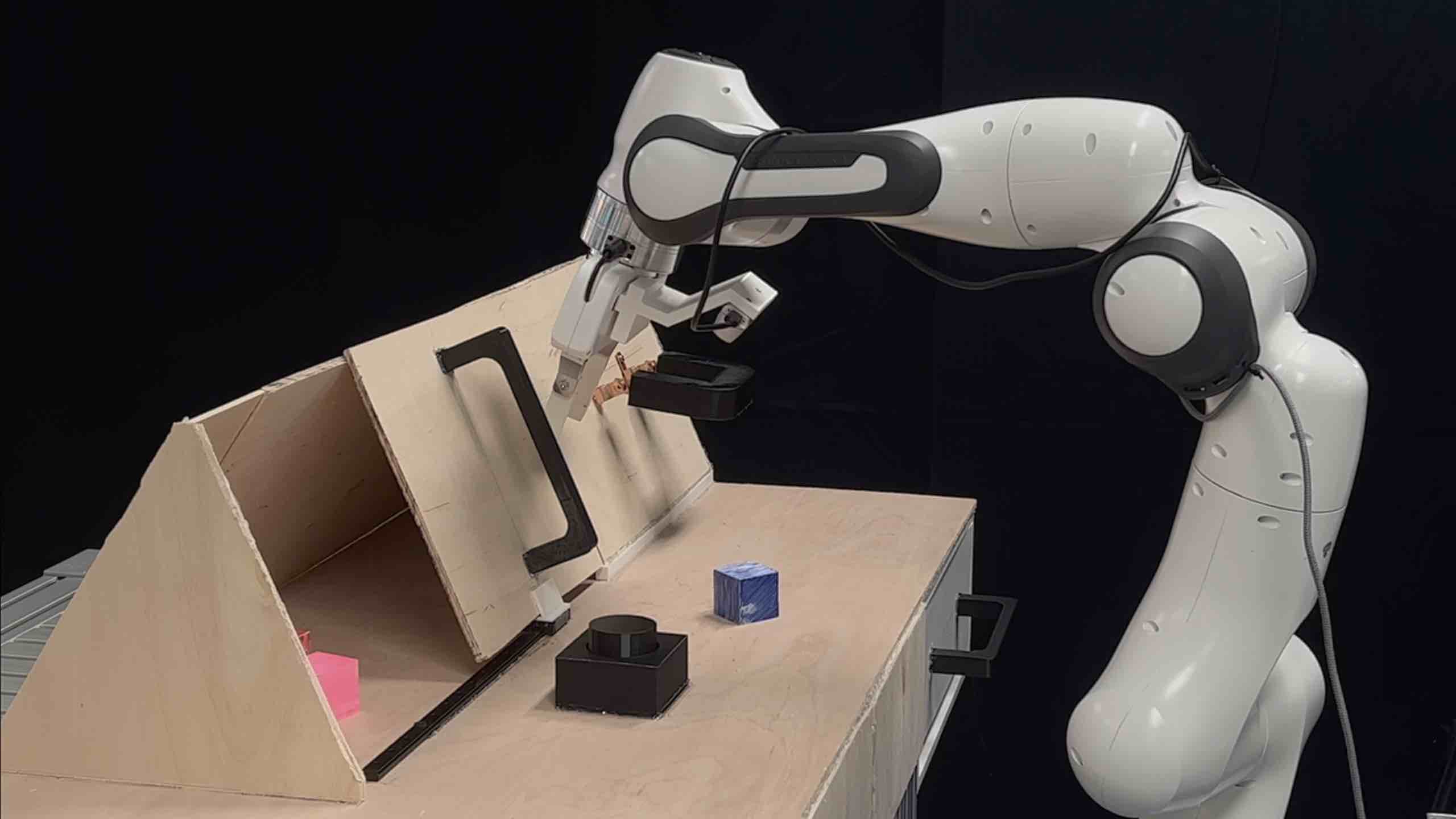}} 
\quad
\subfloat[ move slider right]{\includegraphics[width=.19\textwidth]{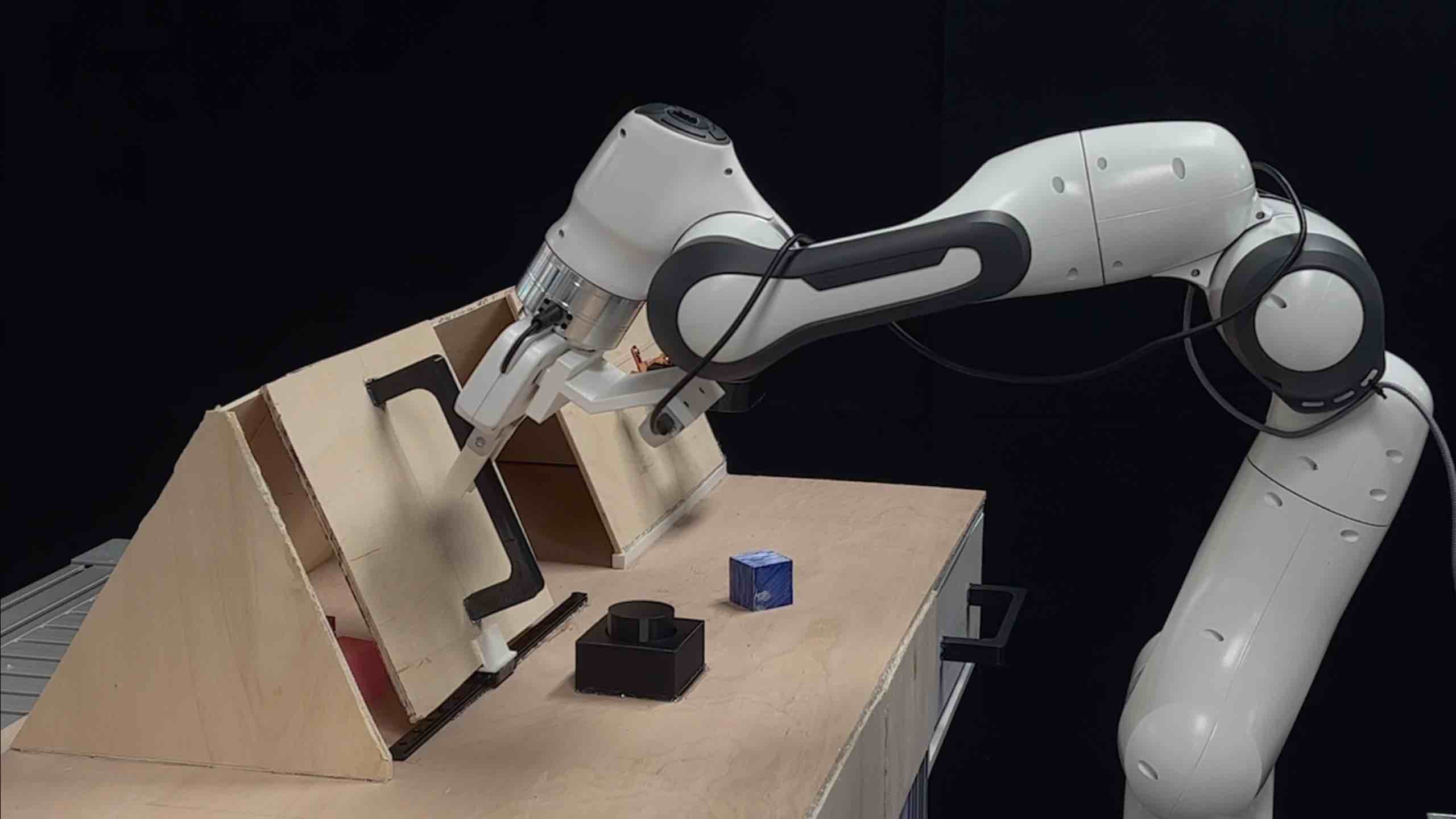}}
\hfill
\subfloat[ open the drawer]{\includegraphics[width=.19\textwidth]{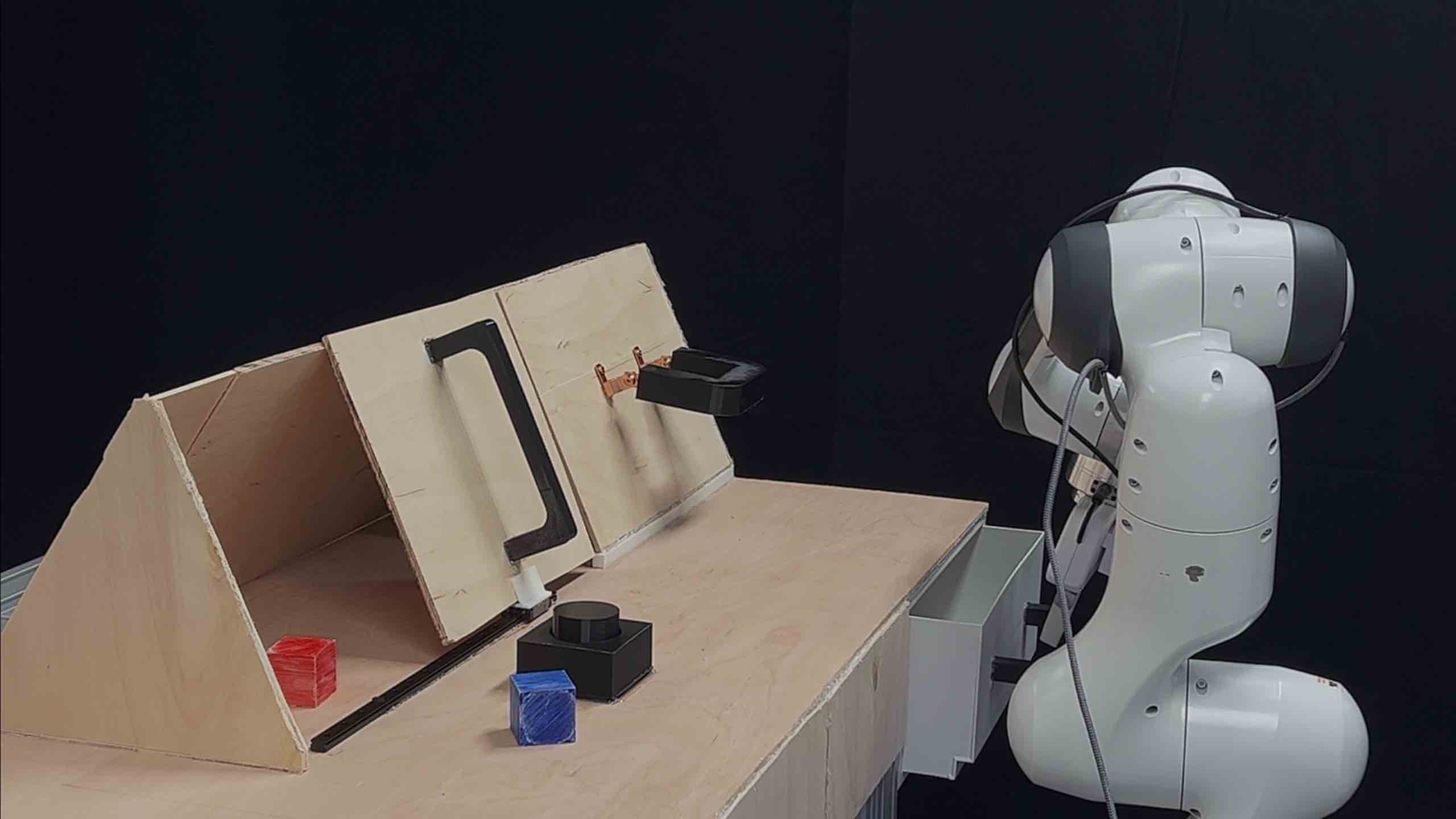}}
\hfill
\subfloat[ close the drawer]{\includegraphics[width=.19\textwidth]{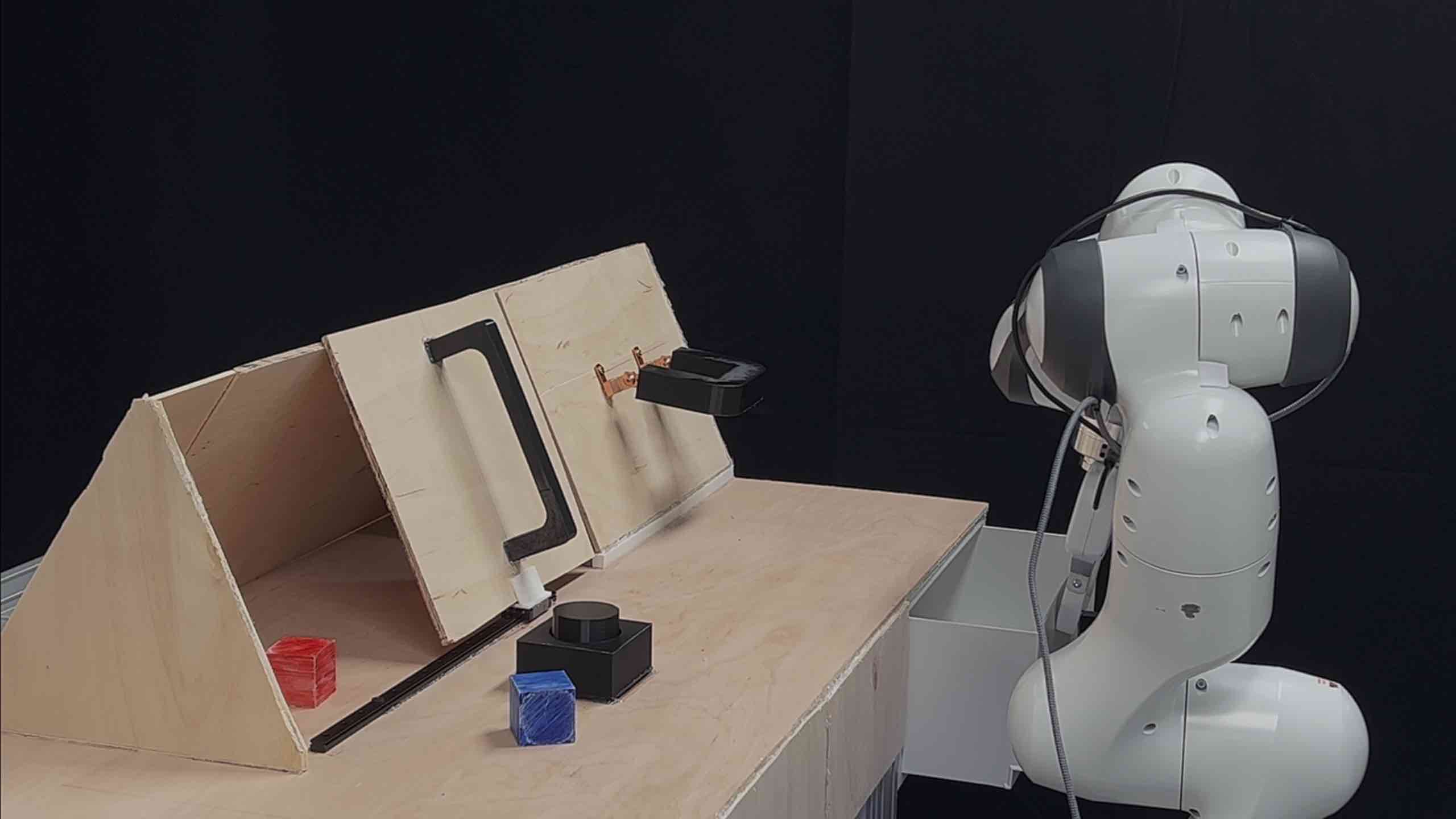}}
\hfill
\subfloat[ toggle switch on]{\includegraphics[width=.19\textwidth]{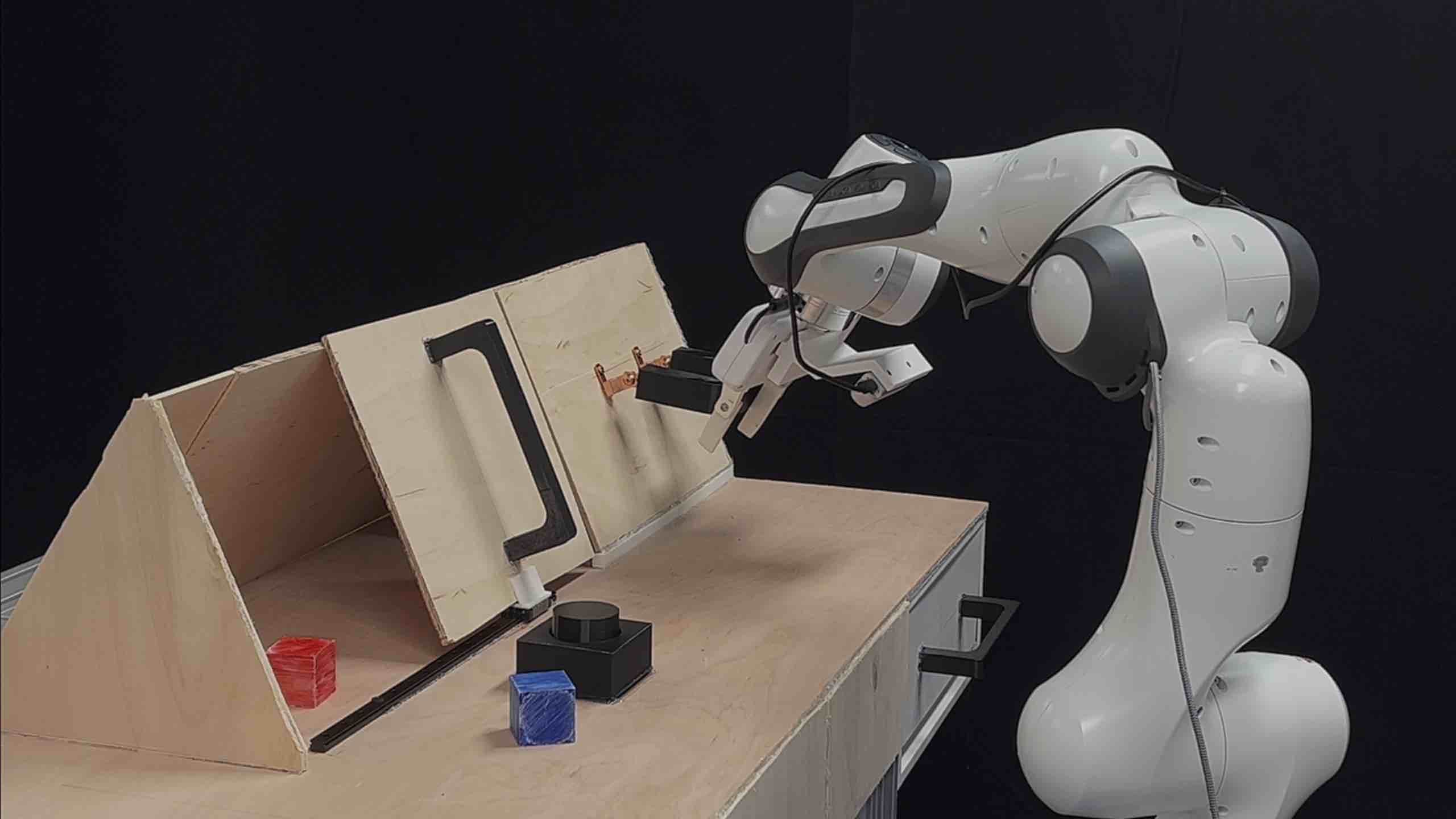}}
\hfill
\subfloat[\scriptsize toggle switch off]{\includegraphics[width=.19\textwidth]{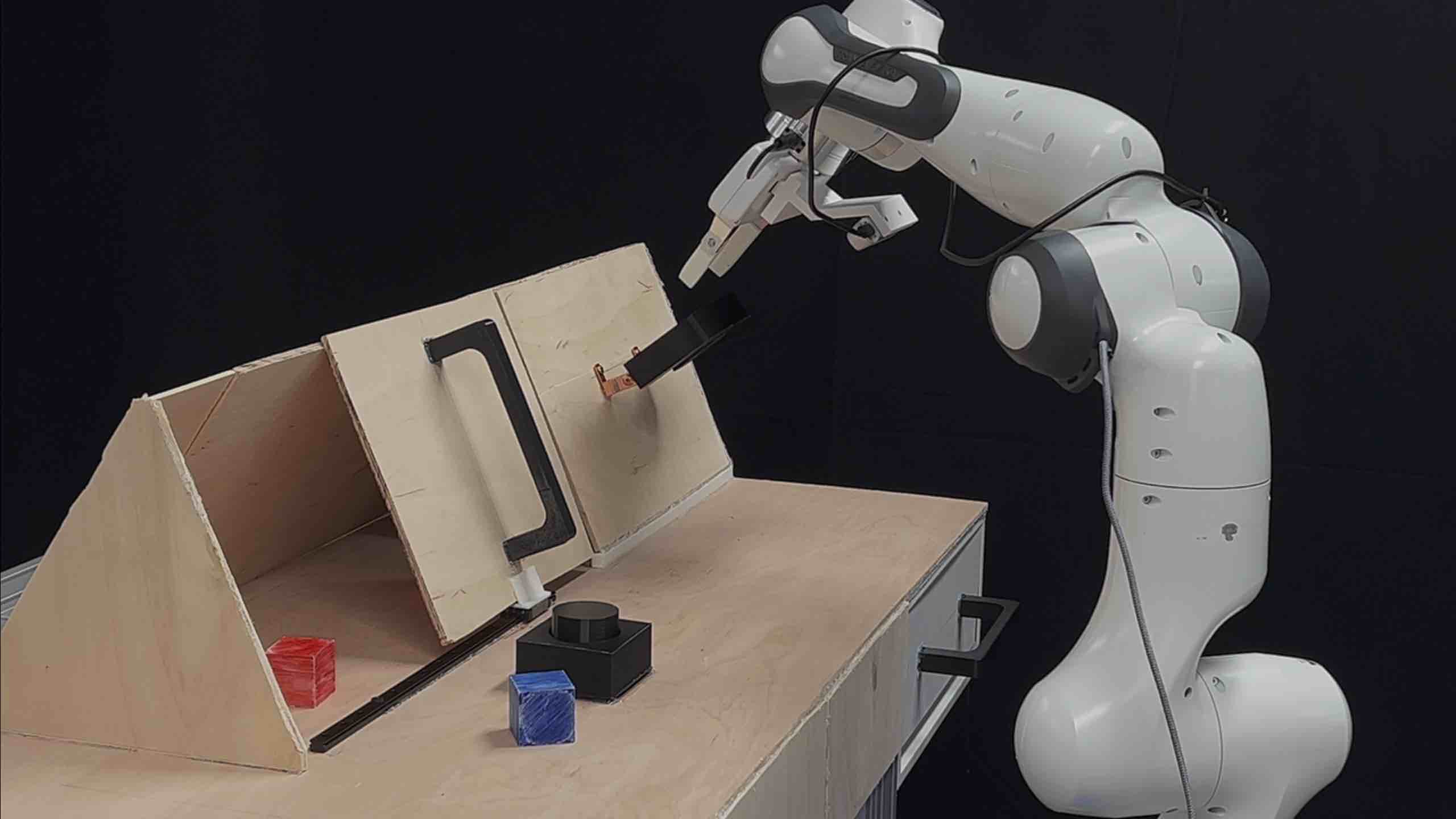}}

\caption{Real-world experiments. We employ the multi-task language control (MTLC) setting in the CALVIN benchmark, encompassing a total of 10 tasks as listed above. The agent is trained in the simulated CALVIN environment D and directly applied to the real-world setting.}

\label{fig:real-world experiment}
\vspace{-1.5em}
\end{figure*}
To investigate the viability of the policy trained in a simulated environment to real-world scenarios, we conduct a sim2real experiment without any additional specific adaptation (\textbf{zero-shot}), as shown in Figure \ref{fig:real-world experiment}.

We designed the real-world environment to closely resemble the simulated CALVIN environment D. The rightmost part of Figure \ref{fig:spil} illustrates that the real-world environment comprises one switch, one cabinet with a slider, one button, one drawer, and three blocks in red, pink, and blue colors. Additionally, two RGB cameras are employed to capture the static observation and gripper observations.

\begin{table}[ht]
    \centering
        \caption{Real-world Experiment Results}
        \resizebox{.48\textwidth}{!}{
        \begin{tabular}{ c c c | c c c } 
            \toprule
            Tasks & HULC & SPIL & Task & HULC & SPIL \\
            \midrule
            open drawer & 0\% & 30\% & move slider right & 0\% & 30\%  \\
            close drawer &  0\% & 40\% & push button & 10\% & 50\%  \\
            toggle switch on & 10\% & 40\% & lift red block & 0\% & 20\%  \\
            toggle switch off & 10\% & 30\% & lift blue block & 0\% & 20\%  \\
            move slider left & 0\% & 40\% & lift pink block & 0\% & 30\%  \\
            \midrule
            \multicolumn{6}{c}{Average: \quad HULC (3\%) \quad SPIL (33\%)} \\
            \bottomrule
        \end{tabular}
        }
        \label{table:real-world-results}
\vspace{-1em}
\end{table}

Table \ref{table:real-world-results} lists the tasks performed and the corresponding success rate. The agent is trained in four CALVIN environments (A, B, C, D), and the trained policy is directly applied to a real-world environment. To mitigate the influence of the robot's initial position on the policies, we execute 10 roll-outs for each task, maintaining identical starting positions. The table results demonstrate our model's effectiveness in handling the challenging zero-shot sim2real experiments. 
Despite the substantial differences between the simulation and real-world contexts, our model still achieves an average success rate of 33\% in accomplishing the tasks. Conversely, the HULC model-trained agent struggles with these tasks, with a  3\% average success rate, underscoring the difficulty of solving real-world challenges. The results from real-world experiments further substantiate our claim that our proposed method exhibits superior generalization capabilities, enabling successful task completion even in unfamiliar environments.

\section{Conclusion}\label{section:conclusion}
In this paper, we introduced a novel imitation learning paradigm that integrates base skills into imitation learning. Our proposed SPIL model effectively improves the generalization ability compared to current baselines and substantially surpasses the SOTA models on the language-conditioned robotic manipulation CALVIN benchmark, especially under the challenging zero-shot multi environment setting. This work also aims to contribute towards the development of general-purpose robots that can effectively integrate human language with their perception and actions.

\bibliographystyle{IEEEtran}
\bibliography{bibliography}

\newpage

 

% \end{IEEEbiography}



\vfill

\end{document}